\documentclass{article}

\usepackage{microtype}
\usepackage{graphicx}
\usepackage{subfigure}
\usepackage{booktabs} 

\usepackage{hyperref}

\usepackage[accepted]{icml2024}

\usepackage{amsmath}
\usepackage{amssymb}
\usepackage{mathtools}
\usepackage{amsthm}
\usepackage{subfigure}

\usepackage{multirow, makecell}

\usepackage[capitalize,noabbrev]{cleveref}

\theoremstyle{plain}

\theoremstyle{definition}

\theoremstyle{remark}

\def\eg{\emph{e.g.}} 
\def\ie{\emph{i.e.}}

\usepackage[textsize=tiny]{todonotes}

\icmltitlerunning{DMTG: One-Shot Differentiable Multi-Task Grouping}

\begin{document}

\twocolumn[
\icmltitle{DMTG: One-Shot Differentiable Multi-Task Grouping}



\icmlsetsymbol{equal}{*}

\begin{icmlauthorlist}
\icmlauthor{Yuan Gao}{cs-whu,ei-whu}
\icmlauthor{Shuguo Jiang}{cs-whu}
\icmlauthor{Moran Li}{tencent}
\icmlauthor{Jin-Gang Yu}{scut}
\icmlauthor{Gui-Song Xia}{cs-whu}
\end{icmlauthorlist}

\icmlaffiliation{ei-whu}{School of EI, Wuhan University}
\icmlaffiliation{cs-whu}{School of CS, Wuhan University}
\icmlaffiliation{tencent}{Tencent Youtu Lab}
\icmlaffiliation{scut}{School of Automation Science and Engineering, South China University of Technology}

\icmlcorrespondingauthor{Gui-Song Xia}{guisong.xia@whu.edu.cn}

\icmlkeywords{Machine Learning, ICML}

\vskip 0.3in
]



\printAffiliationsAndNotice{}  

\begin{abstract}
We aim to address Multi-Task Learning (MTL) with a large number of tasks by Multi-Task Grouping (MTG). Given $N$ tasks, we propose to \emph{simultaneously identify the best task groups from $2^N$ candidates and train the model weights simultaneously in one-shot}, with \emph{the high-order task-affinity fully exploited}. This is distinct from the pioneering methods which sequentially identify the groups and train the model weights, where the group identification often relies on heuristics. 
As a result, our method not only improves the training efficiency, but also mitigates the objective bias introduced by the sequential procedures that potentially lead to a suboptimal solution. Specifically, \emph{we formulate MTG as a fully differentiable pruning problem on an adaptive network architecture determined by an underlying \texttt{Categorical} distribution}. To categorize $N$ tasks into $K$ groups (represented by $K$ encoder branches), we initially set up $KN$ task heads, where each branch connects to all $N$ task heads to exploit the high-order task-affinity. Then, we gradually prune the $KN$ heads down to $N$ by learning a relaxed differentiable \texttt{Categorical} distribution, ensuring that each task is exclusively and uniquely categorized into only one branch. Extensive experiments on CelebA and Taskonomy datasets with detailed ablations show the promising performance and efficiency of our method. The codes are available at https://github.com/ethanygao/DMTG.
\end{abstract}

\begin{figure}[h]
    \centering
    \includegraphics[width=\columnwidth]{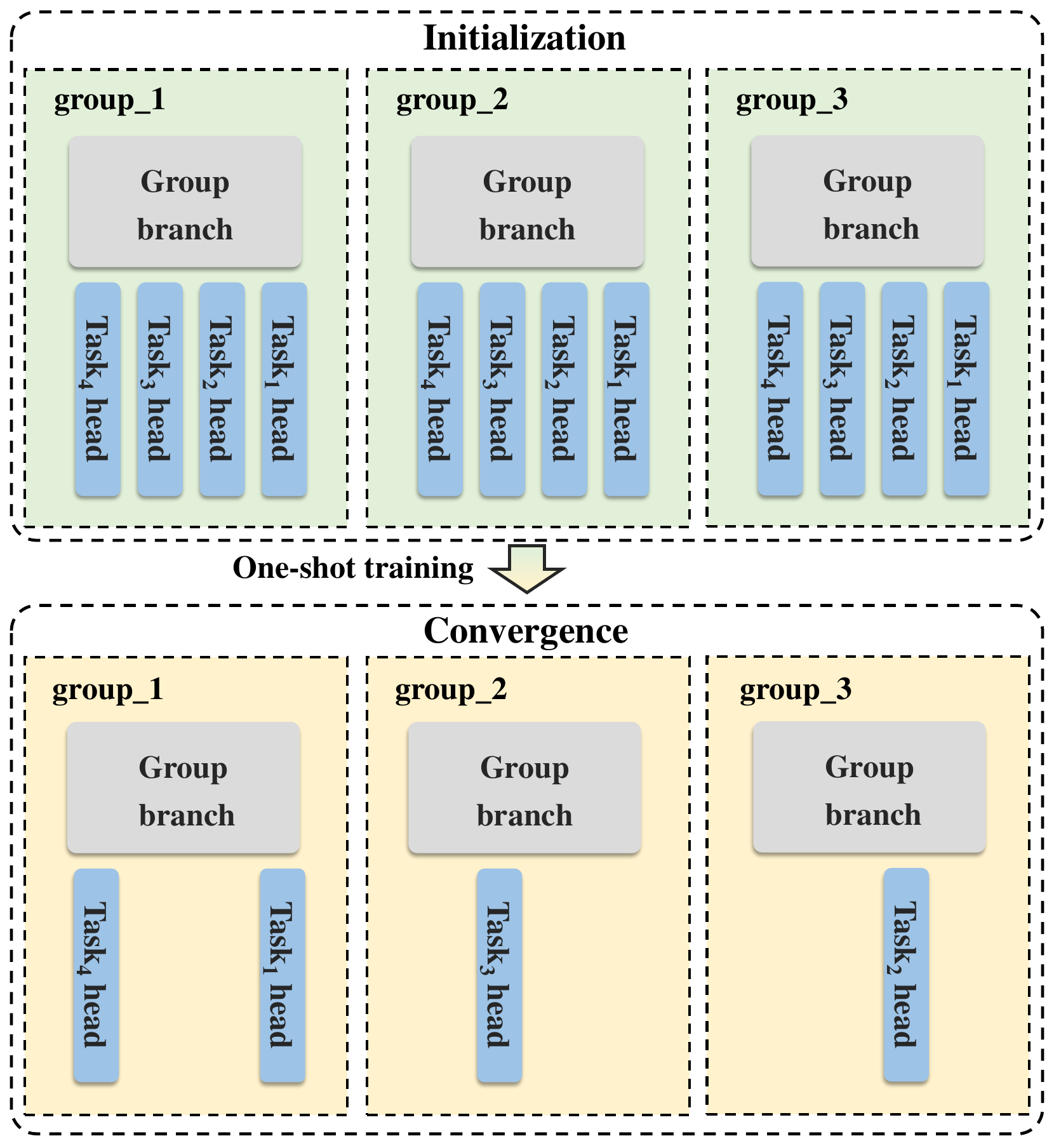}
    \vspace{-8mm}
    \caption{We formulate the Multi-Task Grouping (MTG) problem as network pruning. This figure illustrates the categorization of 4 tasks into 3 groups, where each branch represents a task group. As shown in the Upper Subfigure, at initialization, each group connects to all the task heads, ensuring full exploration of high-order task-affinity. Throughout MTG training, we simultaneously prune the task heads and train the weights of the group-specific branches. Our training process ensures that MTG converges to a categorization where each task exclusively and uniquely belongs to only one group, as illustrated in the Lower Subfigure.}
    \label{img:pruning}
    \vspace{-4mm}
\end{figure}

\vspace{-5mm}
\section{Introduction} \label{sec:intro}
\vspace{-1mm}

Many real-world applications are essentially complex systems that involve the collaboration of a large number of tasks. For example, in autonomous driving \cite{caesarnuscenes2020,huplanning2023}, the system needs to simultaneously perform lane detection \cite{tangreview2021}, depth estimation \cite{godarddigging2019}, vehicle detection and instance segmentation \cite{maskrcnn,maskformer2021}, pedestrians localization \cite{bertonimonoloco2019}, etc. In order to tackle these real-world challenges, it is crucial to simultaneously learn a large number of diverse tasks within a Multi-Task Learning (MTL) framework \cite{ubernet2017,lwr2019,blitznet2017,mtl2016,mtl3d2016,lirobust2021,wanggradient2021,liutowards2020,yugradient2020}, which reduces the inference time and facilitates an improved performance by leveraging the affinity among different tasks.

It is thus critical to harness the affinity among those diverse tasks. Compared to learning them independently, simply combining them and feeding them into a fully shared network oftentimes deteriorates the performance of several or even most tasks. Such phenomenon is attributed to the presence of the inherent \emph{negative transfer}, where the intuition is that the gradients from different tasks may interfere with each other when flowing into a shared encoder.

Pioneering works alleviate the negative transfer by designing novel Multi-Task Architectures (MTA) \cite{mtl1997,convexmtl2008,mtl2011,mtaoverview2017,mtaoverreview2021} or applying the Multi-Task Optimization (MTO) methods \cite{sener2018,lin2019,liuconflict2021,suteuregularizing2019,yangdeep2016}, where state-of-the-art MTA assigns independent network parameters to different tasks, while MTO directly manipulates the gradients from different tasks before applying them to update the shared parameters. However, both MTA and MTO methods pose challenges when scaling to a large number of tasks, \ie, the scalability is impeded in MTA for both training and evaluation due to the extra parameters, while the training of MTO cannot maintain scalability because it has to retain the backward graphs for each task. Recent researches also suggest that it is difficult to address the negative transfer solely by gradient manipulation in MTO \cite{xin2022,kurin2022}.

We instead propose to learn a large number of tasks by Multi-Task Grouping (MTG) \cite{hoa2020}. In MTG, input tasks are categorized into groups by their affinity, where a group of tasks, instead of a single task, is modeled by a unique encoder. When the group categorization is given, MTG for $K$ groups of $N$ tasks drastically reduces the training complexity from $O(N)$ (for MTA/MTO) to $O(K)$. 

The primary challenge of MTG is to identify the group categorization, which involves investigating the exponential $2^N$ group candidates at maximum, given merely $N$ tasks. In order to migrate this issue, Standley et al. \yrcite{hoa2020} and Fifty et al. \yrcite{tag2021} propose to average the pairwise affinities to approximate the high-order affinities\footnote{For example, the performance of Task $A$ in Group $\{A, B, C\}$ is approximated by averaging that of $A$ in $\{A, B\}$ and $\{A, C\}$}. Despite a reduced complexity from $2^N$ to $N^2$, the less precise assumption of linear tasks affinity in \cite{hoa2020,tag2021} degrades the final performance. On the other hand, Song et al. \cite{mtg2022} advocate to train a meta-learner that directly predicts the final performance given a group categorization. However, the training of the meta-learner \emph{per se} is extremely difficult and involves collecting numerous well-trained group samples. Moreover, existing methods perform group identification and grouped task learning in separated sequential procedures. As a result, the former potentially introduces objective bias \emph{w.r.t.} the latter, especially when the groups are categorized based on heuristics. This also leads to potential performance degradation.

In view of those limitations, we propose to \emph{formulate MTG as a pruning problem of an adaptive network architecture}, as shown in Figure \ref{img:pruning}, which enables to 1) \emph{identify the best groups and train the grouped model weights simultaneously in one-shot}, as well as 2) \emph{fully exploiting the high-order task affinities}. In our unified one-shot learning, we formulate the group identification as the model architecture learning/pruning, and the grouped task learning is established as the model weights training under a certain architecture. In this way, both procedures mutually facilitate each other to a better convergence. We jointly train both procedures simply by the task losses, where the high-order task affinities are directly exploited. Our approach excels in both efficiency and accuracy, which is distinct from pioneering two-shot methods that first approximately identify the grouping results, then train the grouped model from scratch subsequently.

Specifically, we formulate the categorization of $N$ tasks into $K$ groups as learning of a \texttt{Categorical} distribution, where the \texttt{Categorical} distribution is used to determine an adaptive network architecture. We then optimize the unified group identification and grouped task learning leveraging a pruning algorithm that is fully differentiable. To this end, our method starts with $K$ branches, each equipped with $N$ heads.
It indicates that at the beginning, all the tasks are predictable by every group, ensuring full exploitation of the high-order task-affinity. After that, we optimize the model weights as well as the \texttt{Categorical} distribution such that the $KN$ heads are gradually pruned down to $N$, facilitating that each task is exclusively and uniquely predicted by only one branch. Our \texttt{Categorical} distribution is continuously relaxed and then optimized by Gumbel softmax \cite{maddison2016concrete}. Our pruning procedure \emph{per se} is efficient, as we only expand the light-weighted \emph{task heads} (\eg, only the last network layers), instead of the heavy \emph{encoders}, and the $K$ encoder branches (each for a group) in our method represent the minimal requirement of MTG.

Our method has been extensively validated on CelebA \cite{faceattribute2015} and Taskonomy \cite{taskonomy2018} with detailed ablations. Our method exhibits two unique features:
\begin{itemize}
\vspace{-2mm}
\item \textbf{Accuracy} with high-order task affinities exploited, which is ensured by 1) the grouping formulation of learning a continuous relaxed and differentiable \texttt{Categorical} distribution, and 2) the elimination of the objective bias by the one-shot training of unified group identification and grouped task learning.
\item \textbf{Efficiency} with $O(K)$ training complexity given $K$ groups, which comes from 1) our pruning formulation instead of sampling group candidates to train from scratch, and 2) the one-shot training that unifies group identification and grouped task learning.
\end{itemize}

\section{Related work}
\label{sec: related work}

\subsection{Multi-Task Grouping}
Multi-Task Grouping (MTG) aims to put collaborative tasks from a task pool into the same group, where a group of tasks can be learned efficiently by a shared network \cite{mtl2011,learningtg2012,jointgroups2021}.
Grouping tasks enables efficient learning of a vast array of tasks while also maintaining high interpretability.
However, the primary challenge in MTG is that finding an optimal grouping solution in $2^N-1$ grouping candidates can be difficult. Existing grouping methods \cite{hoa2020,tag2021,mtg2022} have attempted to model an evaluation function to determine high-order task relationships based on low-order observations. Nonetheless, these methods perform group identification and grouped task learning separately, and potentially considering only low-order task affinity.
In contrast, our grouping approach integrates group identification and grouped task learning within a one-shot training process, significantly improving running efficiency in large-scale task scenarios while thoroughly considering higher-order task relationships.

\subsection{Multi-Task Architecture} 

Multi-Task Architecture (MTA) \cite{mtl1997,convexmtl2008,mtl2011,mtaoverview2017,mtaoverreview2021} is a prevailing technology line in the Multi-Task Learning domain. 
It can be categorized as hard-parameter sharing \cite{ubernet2017,lwr2019,blitznet2017,mtl2016,mtl3d2016} and soft-parameter sharing \cite{nddr2019,gao2020mtl,gao2024aux,crosslingual2015,mrn2017,crossstitch2016}. The former shares a common feature extraction module among tasks, while the latter assigns a special feature extraction branch for each task, exchanging features through extra fusion modules. 
Although great success has been witnessed in designing novel MTL network architectures, they are less appropriate in addressing an extreme large number of tasks. Specifically, it is difficult to avoid the negative transfer due to a full-shared encoder module in hard-parameter sharing methods \cite{vandenhende2019,learntobranch2020,bruggemann2020,adashare2020}, while soft-parameter sharing methods \cite{ruderlatent2019,zhangjoint2018,padnet2018,zhangpattern2019} better address the negative transfer but introduce efficiency issues.

\subsection{Multi-Task Optimization} 
Multi-Task Optimization (MTO) develops in parallel with Multi-Task Architecture, which aims to adjust task loss to balance the learning process of different tasks \cite{uncertainty2018,gradnorm12018,gradnorm22019,weightingtask12022,weightingtask22020,rearranging2018}. Advanced MTO methods directly manipulate gradients from different tasks to mitigate the training conflicts \cite{lirobust2021,wanggradient2021,liutowards2020,yugradient2020}, \eg, projecting task gradients when their angle is greater than $90^{\circ}$. In practice, revising gradients necessitates additional memory to store the gradient graph for each task, which can be potentially infeasible when dealing with an extremely large number of tasks. Most recently, Kurin et al. \yrcite{kurin2022} and Xin et al. \yrcite{xin2022} reveal that the existing MTO methods may be sensitive to hyperparameters when dealing with different combinations of tasks. Our method aims to learn the categorization of tasks and is orthogonal to MTO methods.

\subsection{Network Pruning} 

Network pruning \cite{caiproxylessnas2018,chenreinforced2018,elskenneural2019,ghiasifpn2019,hemilenas2020,lipartial2019} aims to detect and remove the redundancy of the networks without significant performance degradation. This pruning process can be implemented by Bayesian optimization \cite{makingbergstra2013}, evolutionary algorithms \cite{regularizedreal2019,xiegenetic2017}, network transformation \cite{gordonmorphnet2018}, reinforcement learning \cite{irlasguo2019,mnasnettan2019,learningzoph2018}, and gradient descent \cite{akimotoadaptive2019,liuauto2019,wufbnet2019,zhangcustomizable2019}. 
We use differentiated pruning operations, which effectively enable integrating group identification with grouped task learning jointly in one-shot training.
We are the first to implement network pruning into MTG to unify group identification and grouped task learning in an end-to-end architecture.

\vspace{-1mm}
\section{Method}
\label{sec:Method}

\begin{figure*}[t]
\vspace{-1mm}
    \centering
    \includegraphics[width=\textwidth]{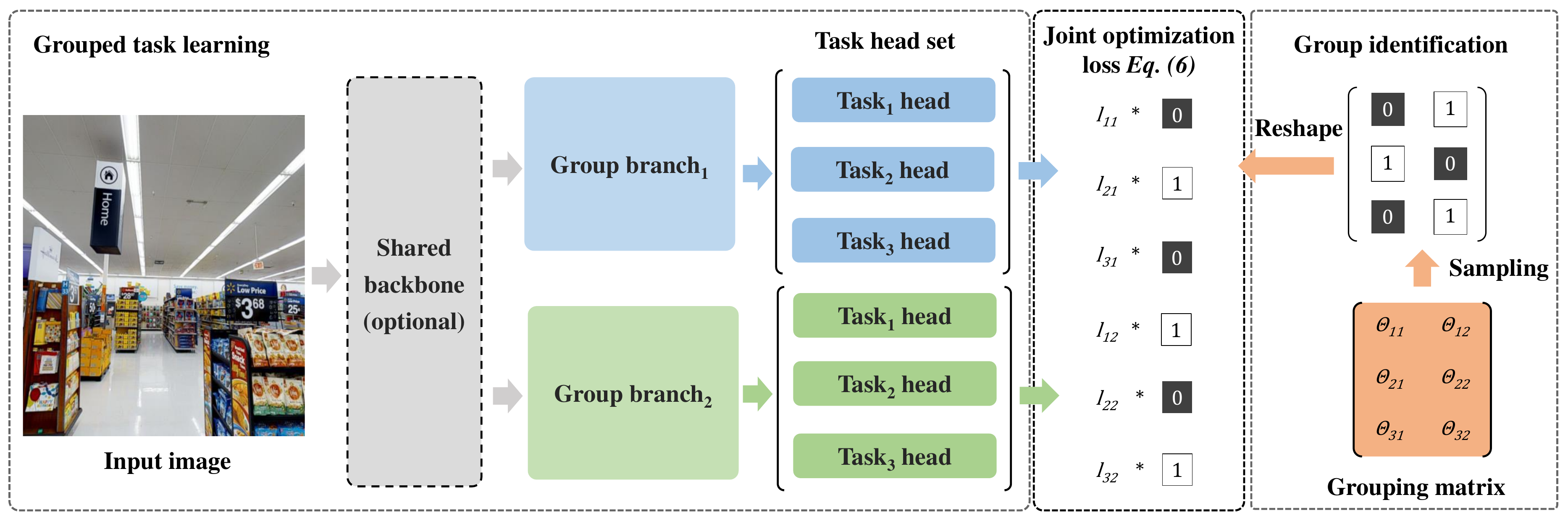}
    \vspace{-8mm}
    \caption{The overview of our method. We formulate the Multi-Task Grouping (MTG) problem as network pruning, where our method consists of a grouped task learning module and a group identification module. In order to categorize $N$ tasks into $K$ groups, our network is constructed with $K$ group-specific branches, optionally with shared lower layers. At initialization, we connect each branch to all the task heads (enabling them to predict all tasks), so that the high-order task-affinity can be exploited. We then formulate the grouped task learning as the model weights training for each group-specific branch, and the group identification as the network head pruning. The final grouped task losses are generated by the element-wise product of both modules, which in turn ensures both modules to be trained simultaneously in one-shot with the high-order task-affinity fully exploited. This figure illustrates categorizing 3 tasks into 2 groups.}
    \label{img:framework}
    \vspace{-2.5mm}
\end{figure*}




Given $N$ tasks, we aim to efficiently chase the best categorization from the $2^N$ possibilities, with the high-order task affinities directly exploited. To this end, we \emph{formulate MTG into a network pruning framework}, where we model the group identification as the architecture learning/pruning, and the grouped task learning as the model weights optimization under a certain architecture. As a result, the group identification and the grouped task learning are unified and can be jointly optimized in one-shot during the network pruning. Regarding the optimization, we design the group categorization as the learning of a \texttt{Categorical} distribution, which is then continuously relaxed into a differentiable \texttt{Concrete} distribution and subsequently optimized using the Gumbel softmax \cite{maddison2016concrete}.


In summary, our method is able to 1) exploit the high-order task affinities directly. 2) It avoids the potential objective bias when group identification and grouped task learning act as separated sequential procedures.
3) Given $K$ groups, our pruning algorithm preserves the efficiency of $O(K)$ training complexity for the encoder. 4) Our \texttt{Categorical} distribution formulation guarantees each task to be categorized into one group \emph{exclusively and uniquely}. Thus, our learned groups and model weights are readily to use without retraining or validation (validation is needed when a certain task is categorized into multiple groups \cite{hoa2020,tag2021,mtg2022}, as discussed in Appendix \ref{appdix:categorization}).

\vspace{-1mm}
\subsection{Problem Formulation}
Formally, we consider categorizing a set of $N$ tasks $\mathcal{T} = \{\mathcal{T}_{1}, ..., \mathcal{T}_{N}\}$ into equal or less than $K$ groups $\mathcal{G} = \{\mathcal{G}_{1}, ..., \mathcal{G}_{K}\}$, such that each group contains 0 to $N$ tasks $\mathcal{G}_k = \{..., \mathcal{T}_{i}, ... \}$, and each task is exclusively and uniquely belongs to one group. Therefore, we have:
\vspace{-1mm}
\begin{align}
    &\mathcal{T} = \cup _{k=1}^{K} \mathcal{G}_{k}, \notag \\
    s.t. \quad &\forall k, \quad |\mathcal{G}_k| \in \{0, ..., N\}, \nonumber \\ 
    &\forall (i, j), \quad \mathcal{G}_{i}\cap \mathcal{G}_{j}=\emptyset,
    \label{eq:grouping}   
\end{align}
where $|\cdot|$ is the cardinality. We optimize our problem exclusively to attain the highest average performance across these $N$ tasks, without relying on heuristic criteria. We also note that $K$ is the maximal-allowed number of groups, and we do not impose a strict requirement to yield precisely $K$ groups, \eg, some groups may contain 0 task.

\textbf{Objective Bias in Two-Stage MTG Methods.} The objective bias in pioneering \cite{hoa2020,tag2021,mtg2022} appears in two aspects: 1) \emph{the group categorization is determined by heuristics but the retraining is based on the optimization of task losses}, and 2) \emph{the difference in the inputs to the group identification and the grouped-model weights retraining stages lead to different objectives}. In other words, the groups are identified heuristically when all the $N$ tasks can synergy/regularize each other, but the retraining phase only sees a subset (group) of tasks thus exhibiting different gradients from the former.

As shown in Figure \ref{img:framework}, the group identification and the grouped task (weights) learning in our method complement each other and are trained jointly in one-shot. On one hand, during the training of the task groups, the group identification module selects collaborative tasks to back-propagate gradients to the corresponding branch of the grouped task learning module. On the other hand, each branch of the grouped task learning module is responsible for one group, which in turn facilitates group identification.


\vspace{-1.5mm}
\subsection{Grouped Task Learning Module} 

We start with $K$ branches in the grouped task learning module, where each branch represents the encoder of each task group. We connect each branch to $N$ task heads to predict all the $N$ tasks, facilitating the exploration of high-order task affinity. Our method possesses an efficient training complexity of $O(K)$ for the network encoder. 

Our method also enables to further reduce the training complexity, by implementing optional group-wise shared layers before splitting into the group-specific branches. This is illustrated in the dashed gray box in Fig. \ref{img:framework}.

\vspace{-1.5mm}
\subsection{Group Identification Module}
\vspace{-0.5mm}

We model the categorization of $N$ tasks exclusively and uniquely into $K$ groups as the learning of an unknown \texttt{Categorical} distribution, where the \texttt{Categorical} distribution is used to determine an adaptive network architecture. As such, the underlying \texttt{Categorical} distribution can be optimized jointly with the model weights in one-shot, which we formulate as a pruning problem.

\noindent \textbf{\texttt{Categorical} Distribution Modeling.} Formally, let a random variable $z_{ik}$ indicate the assignment of task $i$ to group $k$, which is sampled from some discrete distribution. In order to assign $N$ tasks to $K$ groups, we have a set of random variables $Z = \{z_{ik} \} \in \mathbb{R}^{N \times K}$. 

Recall Eq. \eqref{eq:grouping} that each task is exclusively and uniquely categorized into one group, therefore, we have:
\begin{equation}
    \sum_k z_{ik} = 1, \qquad \text{and} \qquad z_{ik} \in \{0, 1\}, \quad \forall k,
    \label{eq:k_row}   
\end{equation}
which indicate that each row of $Z$ follows a \texttt{Categorical} distribution. Let the \texttt{Categorical} random variable $z_{ik}$ be parameterized by $s_{ik}$, we have:
\begin{equation}
    z_{ik} \sim \texttt{Categorical}(s_{ik}),
    \label{eq:cate}   
\end{equation}
where $s_{ik}$ is the probability of assigning task $i$ to group $k$.

\noindent \textbf{Network Architecture Formulation.} We establish an adaptive network architecture by the \texttt{Categorical} distribution, and formulate a pruning problem so that the Group Identification Module and the Grouped Task Learning Module can be optimized jointly in one-shot.

To this end, we formulate the sampled random variable $z_{ik}$ as a loss indicator or a task selector, which determines whether to back-propagate the loss of task $i$ to the $k$-th group. Specifically, let $L = \{L_{ik} \} \in \mathbb{R}^{N \times K}$ be the loss matrix of $KN$ task heads, the final loss can be obtained by:
\begin{equation}
    L^{\text{task}}(\theta, S) = L(\theta) \odot Z(S),
    \label{eq:loss}
\end{equation}
where $\odot$ is the element-wise product, $\theta$ is the model weights, and $S = \{ s_{ik} \} \in \mathbb{R}^{N \times K}$ is the set of parameters of the \texttt{Categorical} distributions.

As shown in Eq. \ref{eq:loss}, we formulate MTG as a pruning problem where $Z(S)$ is learned to prune the $KN$ losses $L(\theta)$. We note that the cost of training a $N \times K$ matrix $S$ and sampling $Z$ from $S$ is negligible \emph{w.r.t.} the learning of $K$ group encoders, retaining the training complexity of our pruning formulation as $O(K)$ for the heavy encoder\footnote{Our training efficiency can be impeded in uncommon cases when the network heads are heavy, we discuss this in Appendix \ref{appeix:time}.}.

\vspace{-1mm}
\subsection{The Joint Optimization}
\vspace{-1mm}

Equation \eqref{eq:cate} involves a discrete sampling from $s_{ik}$ to $z_{ik}$, which results in a gradient blockage in Eq. \eqref{eq:loss} when back-propagating gradients from $Z$ to $S$. In this section, we continuously relax the discrete \texttt{Categorical} distribution, so that both the parameters for group identification $S$ and the weights for grouped task learning $\theta$ can be jointly optimized in one-shot by back-propagating the gradients from the task loss $L^{\text{task}}(\theta, S)$, through $Z(S)$ and $L(\theta)$, respectively.

\noindent \textbf{Continuous Relaxation.} By using the reparameterization trick from the \texttt{Concrete} distribution \cite{maddison2016concrete}, we are able to continuously sample $s_{ik}$ to produce $\tilde{z}_{ik}$ that approximate $z_{ik}$ of the \texttt{Categorical} distribution. This facilitates the gradient flow from $L^{\text{task}}(\theta, S)$ to $s_{ik}$ through $\tilde{z}_{ik}$. The reparameterized \texttt{Categorical} distribution is modeled by the differentiable Gumbel softmax:
\begin{equation}
    \tilde{z}_{ik} = \frac{\exp((s_{ik}+g_{ik})/\tau)}{\sum _{m=1}^{K}\exp((s_{im}+g_{im})/\tau)}
    \label{eq:gumbel}
\end{equation}
where $g_{ik}$ is sampled from a \texttt{Gumbel} distribution, \ie, $g_{ik}=-\log(-\log(\texttt{Uniform}(0, 1)))$ \cite{maddison2016concrete}. $\tau$ is a small or annealing temperature, producing a discrete $\tilde{z}_{ik}$ after convergence as a good approximation of $z_{ik}$. Given $\widetilde{Z} = \{ \tilde{z}_{ik} \} \in \mathbb{R}^{N \times K}$, the loss in Eq. \eqref{eq:loss} becomes:
\begin{equation}
    L^{\text{relaxed task}}(\theta, S) = L(\theta) \odot \widetilde{Z}(S),
    \label{eq:relaxed_loss}
\end{equation}

\noindent \textbf{Initialization.} We note that the parameter of the \texttt{Categorical} distribution, $s_{ik}$, can be initialized according to the prior knowledge of the task affinity. In our problem, we simply initialize each $s_{ik}$ to $1/K$, which implies that each task has an equal probability of being categorized into any group. In other words, we do not assume any task affinities and learn them in a fully data-driven manner. Based on that, we optimize our model by pruning the initial $KN$ task heads to $N$, where each task is exclusively and uniquely categorized into one group after convergence.

\section{Experiments}
\label{sec:Experiments}
In this section, we extensively validate our method on both \textbf{Taskonomy} \cite{taskonomy2018} and \textbf{CelebA} \cite{faceattribute2015} datasets for various candidate groups. We detail the experimental setup in the following.

\subsection{Experimental Setup} \label{sec:setup}

\noindent \textbf{Datasets.} We perform experiments on the Taskonomy dataset \cite{taskonomy2018} following \cite{hoa2020,tag2021,mtg2022}, and the CelebA dataset \cite{faceattribute2015} following \cite{tag2021}. We use the official tiny train, validation, and test split of Taskonomy. The images from Taskonomy and CelebA are bilinearly downsampled to $256\times 256$ and $64\times64$, respectively. Those datasets are introduced in detail in Appendix \ref{appdix:dataset}.

\noindent \textbf{Benchmark Experiments.} We follow the experiment setups in \cite{hoa2020,tag2021,mtg2022} to conduct $5$ tasks on Taskonomy, \ie, \textit{semantic segmentation}, \textit{depth estimation}, \textit{surface normal}, \textit{keypoint detection}, and \textit{edge detection}, denoted as \textbf{Taskonomy-5}. We also conduct $9$ tasks on CelebA dataset following \cite{tag2021}, \ie, \textit{5\_o\_Clock\_Shadow}, \textit{Black\_Hair}, \textit{Blond\_Hair}, \textit{Brown\_Hair}, \textit{Goatee}, \textit{Mustache}, \textit{No\_Beard}, \textit{Rosy\_Cheeks}, and \textit{Wearing\_Hat}, referred to as \textbf{CelebA-9}. We perform the full 40 tasks of the CelebA dataset, \ie, \textbf{CelebA-40}, in Appendix \ref{appdix:scalability}, showcasing our scalability to numerous tasks.

\noindent \textbf{Network Backbone.} We use the same network backbone as \cite{hoa2020,tag2021}, \ie, a variant of Xception network \cite{xception2017} with 12 blocks, for the Taskonomy experiments. For CelebA, we use a variant of ResNet \cite{resnet2016} following \cite{tag2021}.

\noindent \textbf{Optimization.} We use Adam optimizer for all of our experiments, where the initial learning rates are $0.0008$ and $0.0001$ for the CelebA and Taskonomy experiments, respectively. We use \emph{plateau} learning rate decay which reduces by $0.5$ when the validation loss no longer improves. We train all the experiments for $100$ epochs, where our networks are initialized by the pre-trained naive MTL weights on the corresponding experiments. We copy the networks and the group-specific parameters for $K$ times to ensure that the same task is initialized identically across different groups. We initialize the Gumbel Softmax temperature $\tau$ of Eq. \eqref{eq:gumbel} as $2.5$ and $4$ for the CelebA and Taskonomy experiments, respectively. We follow \cite{tag2021} to use the cross-entropy loss for the CelebA experiments, and follow \cite{hoa2020,tag2021} to use the cross-entropy loss for semantic segmentation and $\ell_{1}$ loss for other tasks of the Taskonomy experiments.

\noindent \textbf{Evaluation Metrics.} Pioneering research in MTG commonly relied on the \emph{total loss} as the evaluation metric \cite{hoa2020,tag2021,mtg2022}, which straightforwardly sums up the losses of all tasks. However, the magnitudes of losses from different tasks significantly vary due to 1) different loss types, such as cross-entropy losses for classification and $\ell_{1}$ losses for regression, and 2) diverse labels, such as image-level classification labels and pixel-level semantic segmentation labels. Consequently, simply calculating the \emph{total loss} may lead to an overestimation of tasks with higher loss magnitudes while overshadowing those with lower loss magnitudes. This phenomenon contradicts the goal of MTG, \ie, boosting \emph{all the input tasks rather than a subset of them} \cite{hoa2020}.

To comprehensively assess improvements of an MTG method across all tasks, we follow \cite{maninis2019attentive,vandenhende2020mti,mtaoverreview2021} to eliminate the influence of loss magnitudes, which is termed \textbf{normalized gain}. Specifically, we initially calculate the normalized loss improvement (expressed as a percentage) \emph{w.r.t.} the naive MTL architecture (\ie, the shared-encoder architecture, oftentimes the worst baseline) for each task, then average them for all tasks:
\begin{equation}
    \text{NormGain}_{L} = \frac{1}{N}\sum_{n=1}^{N}\frac{L^{\text{task } n}_\text{Naive MTL}-L^{\text{task } n}_\text{method}}{L^{\text{task } n}_\text{Naive MTL}},
    \label{eq:taskonomy error}
\end{equation}
where $L$ denotes loss and $N$ is the total number of tasks. Similarly, in cases where a unified evaluation is applicable for all input tasks (\eg, classification error when all input tasks are classifications), we can also present \textbf{normalized gain} \emph{w.r.t.} such unified evaluation error:
\begin{align}
    &\text{NormGain}_{E} = \frac{1}{N} \sum_{n=1}^{N} \frac{E^{\text{task } n}_\text{Naive MTL} - E^{\text{task } n}_\text{method}}{E^{\text{task } n}_\text{Naive MTL}}, 
    \label{eq:classification error}
\end{align}
where $E$ denotes the unified evaluation error.

\begin{table*}[t]
\centering
\resizebox{0.99\textwidth}{!}{
    \begin{tabular}{l|l||cc|cc|cc|cc|cc}
    \hline
     \makecell[c]{\multirow{2}{*}{Groups}}&\makecell[c]{\multirow{2}{*}{Methods}}& \multicolumn{2}{c|}{Depth Estimation} & \multicolumn{2}{c|}{Surface Normal} & \multicolumn{2}{c|}{Semantic Segmentation} & \multicolumn{2}{c|}{Keypoint Detection} & \multicolumn{2}{c}{Edge Detection}\\ \cline{3-12}
     &&Loss $\downarrow$&NormGain$_L$ (\%) $\uparrow$&Loss $\downarrow$&NormGain$_L$ (\%) $\uparrow$&Loss $\downarrow$&NormGain$_L$ (\%) $\uparrow$&Loss $\downarrow$&NormGain$_L$ (\%) $\uparrow$&Loss $\downarrow$&NormGain$_L$ (\%) $\uparrow$ \\ \hline \hline
     \makecell[c]{-}&Naive MTL&8.67e-3&-&	1.07e-1&-&	8.28e-2&-&	1.19e-2&-&	1.31e-2&- \\
     \makecell[c]{-}&STL&1.60e-5&+99.82&	1.07e-1&-0.18&	9.16e-2&-10.63&	1.30e-4&+98.91&	1.56e-4&+98.81 \\ \hline
    \multirow{5}{*}{$K=3$} & RG      &2.57e-2& -195.88      &1.08e-1& -0.81    &8.43e-2& -1.88         &6.87e-3& +42.46        &6.88e-3& +47.45\\
                         & HOA    &5.85e-3& +32.47      &1.11e-1& -4.37   &7.33e-2& +11.49            &\textbf{2.00e-6}& \textbf{+99.98}        &8.60e-5& +99.34 \\
                         & TAG    &5.15e-3& +40.59      &1.21e-1& -12.93   &8.43e-2& -1.88          &\textbf{2.00e-6}& \textbf{+99.98}      &8.60e-5& +99.34\\
                         & MTG-Net &2.04e-4& +97.65      &\textbf{1.07e-1}& \textbf{+0.00}              &8.28e-2& +0.00                      &6.39e-4& +94.65        &4.08e-4& +96.88\\
                         & Ours    &\textbf{1.19e-7}& \textbf{+100.00}                &1.07e-1& -0.05   &\textbf{6.65e-2}& \textbf{+19.64}           &4.30e-5& +99.63        &\textbf{3.58e-7}& \textbf{+100.00}\\ \hline
    \multirow{5}{*}{$K=4$} & RG      &5.15e-3& +40.59         &1.07e-1& +0.00              &7.33e-2& +11.49              &1.19e-2& +0.00                  &6.88e-3& +47.45 \\
                         & HOA    &5.15e-3& +40.59         &1.06e-1& +0.44       &8.33e-2& -0.61              &\textbf{2.00e-6}& \textbf{+99.98}           &8.60e-5& +99.34\\
                         & TAG    &5.15e-3& +40.59         &1.11e-1& -4.37       &7.33e-2& +11.49              &\textbf{2.00e-6}& \textbf{+99.98}           &8.60e-5& +99.34\\
                         & MTG-Net &2.04e-4& +97.65         &1.07e-1& +0.00              &8.28e-2& +0.00                     &6.39e-4& +94.65           &4.08e-4& +96.88\\
                         & Ours    &\textbf{1.19e-7}& \textbf{+100.00}                &\textbf{1.05e-1}& \textbf{+1.43}       &\textbf{6.46e-2}& \textbf{+21.96}              &4.70e-5& +99.61          &\textbf{1.20e-5}& \textbf{+99.91}\\ \hline
    \multirow{5}{*}{$K=5$} & RG      &5.15e-3& +40.59         &1.07e-1& +0.00              &7.33e-2& +11.49              &6.87e-3& +42.46           &6.88e-3& +47.45\\
                         & HOA    &5.15e-3& +40.59         &1.06e-1& +0.44       &8.33e-2& -0.61              &2.00e-6& +99.98           &8.60e-5& +99.34\\
                         & TAG    &5.15e-3& +40.59         &1.11e-1& -4.37       &7.33e-2& +11.49              &2.00e-6& +99.98           &8.60e-5& +99.34\\
                         & MTG-Net &2.04e-4& +97.65         &1.07e-1& +0.00              &8.28e-2& +0.00                     &6.39e-4& +94.65           &4.08e-4& +96.88\\
                         & Ours    &\textbf{1.19e-7}& \textbf{+100.00}                &\textbf{1.05e-1}& \textbf{+1.62}       &\textbf{6.34e-2}& \textbf{+23.43}              &\textbf{1.00e-6}& \textbf{+99.99}           &\textbf{4.17e-7}& \textbf{+100.00}\\ \hline \hline
    \end{tabular}}
    \setcounter{table}{1}
    \vspace{-2mm}
    \caption{Performance (loss) on Taskonomy-5 \emph{w.r.t.} each input task. Other parameters are the same as those in Table \ref{tab:taskonomy2}.}
    \vspace{-4mm}
    \label{tab:taskonomy_per_task}
\end{table*}

\subsection{Experiments on Taskonomy with 5 Tasks}
Following \cite{hoa2020,tag2021,mtg2022}, we compare our methods with the state-of-the-art MTG methods including \textbf{HOA} \cite{hoa2020}, \textbf{TAG} \cite{tag2021}, and \textbf{MTG-Net} \cite{mtg2022}. 

We also report the performance of Random Group (\textbf{RG}) 
which randomly divides the input tasks into a specific number of groups. 
We illustrate the baseline performance with \textbf{Naive MTL}, where all the tasks are trained simultaneously with a fully-shared encoder (\ie, within 1 group). The performance where each task is trained separately without grouping is denoted as Single Task Learning (\textbf{STL}). We perform candidate numbers of groups as 3, 4, and 5.

The results in terms of losses are shown in Table \ref{tab:taskonomy2}. Our method outperforms SOTA methods by a large margin with a more efficient $O(K)$ training complexity for the encoder, we give detailed training time in Appendix \ref{appeix:time}. Our method reduces the total loss by $22\%$ compared to naive MTL and by $13\%$ compared to STL when $K=3$. As $K$ increases, our grouping performance further improves. Regarding the normalized metric NormGain \emph{w.r.t.} loss, \ie, Eq. \eqref{eq:taskonomy error}, it also achieves a remarkable improvement of over $60\%$ \emph{w.r.t.} naive MTL. Consistent observation can be obtained regarding the error statistics (\ie, Eq. \eqref{eq:classification error}), as shown in Appendix \ref{appdix:error_taskonomy}.

\begin{table}[t]
\centering
\resizebox{\columnwidth}{!}{
\begin{tabular}{c|l||cc||c}
\hline
 Groups&\makecell[c]{Methods}&Total Loss $\downarrow$& NormGain$_L$ (\%)  $\uparrow$&Relative Encoder Complex. \\ \hline \hline
 -&Naive MTL&0.223&-&$1$ \\
 -&STL&0.199&+57.35&$O(N)$ \\ \hline
 \multirow{5}{*}{$K=3$}&RG&0.231&-21.73&$O(K)$\\
 &HOA &0.190&+47.78&$O(N^2)+O(K)$\\
 &TAG &0.210&+45.02&$O(N)+O(K)$\\
 &MTG-Net &0.191&+57.83&$-$\\
 &Ours&\textbf{0.173}&\textbf{+63.85}&$O(K)$\\ \hline
 \multirow{5}{*}{$K=4$}&RG&0.204&+19.90&$O(K)$\\
 &HOA &0.195&+47.95&$O(N^2)+O(K)$\\
 &TAG &0.190&+49.41&$O(N)+O(K)$\\
 &MTG-Net &0.191&+57.83&$-$\\
 &Ours&\textbf{0.170}&\textbf{+64.58}&$O(K)$\\ \hline
 \multirow{5}{*}{$K=5$}&RG&0.198&+28.40&$O(K)$\\
 &HOA &0.195&+47.95&$O(N^2)+O(K)$\\
 &TAG &0.190&+49.41&$O(N)+O(K)$\\
 &MTG-Net &0.191&+57.83&$-$\\
 &Ours&\textbf{0.168}&\textbf{+65.01}&$O(K)$\\ \hline \hline
\end{tabular}}
\setcounter{table}{0}
\vspace{-3.5mm}
\caption{Experimental results on Taskonomy-5. We report the total loss and NormGain$_L$. NormGain$_L$ is the normalized gain according to the task loss calculated by Eq. \eqref{eq:taskonomy error}. NormGain$_L$ and relative encoder complexity are measured \emph{w.r.t.} Naive MTL.}
\vspace{-5.5mm}
\label{tab:taskonomy2}
\end{table}

It can be observed that the performances in terms of total loss and NormGain are not consistent for some MTG methods. For example, in Table \ref{tab:taskonomy2} at $K=3$, the \textbf{NormGain} of MTG-Net is over 10\% higher than that of HOA, given that their \textbf{total losses} are comparative. This is because, as discussed in Evaluation Metrics of Sect. \ref{sec:setup}, the total loss is affected by loss magnitudes associated with different tasks, a slight improvement in a task with a large loss magnitude might overshadow a significant degradation in a task with a small loss magnitude. In contrast, the NormGain metrics address this issue by eliminating such undesirable influence through normalization, providing a more reasonable measurement \emph{w.r.t.} the improvement of all the tasks. We further validate this by illustrating the loss and the relative gain \emph{w.r.t.} Naive MTL \emph{for each task} in Table \ref{tab:taskonomy_per_task}, where our method achieves the best performance across almost all tasks.

We also show the training complexity of the heavy encoder relative to the Naive MTL method in Table \ref{tab:taskonomy_per_task}. Given $K << N$, Our method achieves the best training efficiency except for Naive MTL, but Naive MTL fails to deliver desirable accuracy through a fully shared encoder across all the tasks. Note that there are two terms for the training complexity of HOA and TAG, as they involve first identifying task groups according to $N$ tasks, then training $K$ networks, each for a task group, from scratch. The training complexity of MTG-Net is not included as it requires up to $O(2^N)+O(K)$ to sample up to $2^N$ task combinations and subsequent training them from scratch\footnote{For the Taskonomy-5 experiment, MTG-Net exhaustively samples 31 task combinations with 5 input tasks, as reported in \cite{mtg2022}, resulting in a complexity of $O(2^N)+O(K)$. \label{fn:MTG-Net}}. Note that as our method expands $N$ heads to $KN$, our efficiency can be impeded in uncommon cases where the network heads dominate the computations of the whole network, we discuss this in Appendix \ref{appeix:time}.  

\begin{table}[t]
\centering
\resizebox{0.99\columnwidth}{!}{
\begin{tabular}{c|l||cc||c}
\hline
 Groups&\makecell[c]{Methods}&Total Error $\downarrow$&NormGain$_E$ (\%) $\uparrow$&Relative Encoder Complex. \\ \hline \hline
 -&Naive MTL&56.13&-& 1 \\
 -&STL&59.93&-8.70&$O(N)$ \\ \hline
 \multirow{4}{*}[-0.5ex]{$K=2$}
 &RG&54.87&+1.06&$O(K)$ \\
 &HOA &53.60&+3.27&$O(N^2) + O(K)$ \\
 &TAG &53.41&+4.38&$O(N) + O(K)$ \\
 &Ours&\textbf{52.97}&\textbf{+5.75}&$O(K)$ \\ 
 \hline
 \multirow{4}{*}[-0.5ex]{$K=3$}
 &RG&54.57&+1.54&$O(K)$ \\
 &HOA &54.04&+3.62&$O(N^2) + O(K)$ \\
 &TAG &54.37&+2.08&$O(N) + O(K)$ \\
 &Ours&\textbf{53.67}&\textbf{+4.64}&$O(K)$ \\ 
 \hline
 \multirow{4}{*}[-0.5ex]{$K=4$}
 &RG&54.57&+1.54&$O(K)$ \\
 &HOA &54.14&+2.53&$O(N^2) + O(K)$ \\
 &TAG &54.11&+3.17&$O(N) + O(K)$ \\
 &Ours&\textbf{53.62}&\textbf{+4.62}&$O(K)$ \\ 
 \hline \hline
\end{tabular}}
\setcounter{table}{2}
\vspace{-3.5mm}
\caption{Experimental results on CelebA-9. As all the input tasks are classifications, we report the total classification error and NormGain$_E$, \ie, the normalized gain according to the classification error calculated by Eq. \eqref{eq:classification error}. NormGain$_E$ and relative encoder complexity are measured in terms of Naive MTL.}
\vspace{-5.5mm}
\label{tab:celeba}
\end{table}

\vspace{-1mm}
\subsection{Experiments on CelebA with 9 Tasks}

We compare our method with the state-of-the-art methods \textbf{HOA} \cite{hoa2020} and \textbf{TAG} \cite{tag2021}. \textbf{MTG-Net} \cite{mtg2022} is not included in the CelebA-9 experiments as MTG-Net does not scale well \emph{w.r.t.} number of input tasks $N$, \ie, MTG-Net requires to inefficiently sample up to $2^N$ task combinations and subsequent training them from scratch, which may take \emph{thousands of GPU hours} as reported in \cite{mtg2022}.

Following \cite{tag2021}, we perform candidate numbers of groups as 2, 3, and 4 on CelebA-9. As all the input tasks in this experiment are classification tasks, therefore we report the total classification error and NormGain \emph{w.r.t.} classification error, \ie, Eq. \eqref{eq:classification error}, in Table \ref{tab:celeba}. Table \ref{tab:celeba} illustrates the experiment results on CelebA-9, showing that
our method consistently outperforms the state-of-the-art methods on the CelebA-9 experiments by a large margin with a more efficient training complexity of $O(K)$ for the encoder.

\vspace{-2mm}
\section{Ablation Analysis}
\vspace{-1mm}
We carefully investigate the following issues by ablation. 1) Whether our proposed one-shot MTG outperforms the common practice of two-shot methods \cite{hoa2020,tag2021,mtg2022} given the same group categorization in Sect. \ref{sec:one-shot}. 2) Can our method generalize to the transformer backbones in Sect. \ref{sec:transformer}. 3) The flexibility if we share more or less encoder layers in our method in Sect. \ref{sec:sharing}. 4) Can our method scale to more input tasks in Appendix \ref{appdix:scalability}. 5) The influence of different Gumbel Softmax temperatures in Appendix \ref{appeix:tau}. 6) The group categorization identified by different methods in Appendix \ref{appdix:categorization}.


\vspace{-1mm}
\subsection{Merits of One-shot Nature for MTG \label{sec:one-shot}} 

The one-shot simultaneous group identification and grouped task learning is one of the key features of our method. Benefit from that, our method is able to 1) avoid the potential objective bias in the existing two-shot methods \cite{hoa2020,tag2021,mtg2022}, where group identification and grouped task learning act as separated sequential procedures, 2) further accelerate the training procedure as retraining from scratch is no longer needed.

In order to validate those benefits, we perform ablation on Taskonomy-5 using the group categorizations discovered by our method, and compare our method \emph{w.r.t.} two double-shot methods, which are retrained from \emph{scratch} and from \emph{Naive MTL initialization}, respectively. As shown in Table \ref{tab:taskonomy_retrain}, our one-shot method significantly outperforms both two-shot counterparts with the same group categorization. This suggests that the one-shot training strategy of our methods allows \emph{the group branches to see more tasks at the early training stage than they have to predict in the end}, therefore the results of our one-shot method are better than those of training those grouped subsets of tasks from scratch.

\begin{table}[t]
\centering
\resizebox{\columnwidth}{!}{
\begin{tabular}{c|l|cc}
\hline
 Groups&\makecell[c]{Methods}&Total Loss $\downarrow$&NormGain$_L$ (\%) $\uparrow$ \\ \hline \hline
 \multirow{3}{*}{$K=3$}&Retrain from Scratch&0.183&+53.34 \\
 &Retrain from Naive MTL Init.&0.194&+57.10 \\
 &Ours (one-shot)&\textbf{0.173}&\textbf{+63.85} \\ \hline
 \multirow{3}{*}{$K=4$}&Retrain from Scratch&0.183&+53.34 \\
 &Retrain from Naive MTL Init.&0.194&+57.10 \\
 &Ours (one-shot)&\textbf{0.170}&\textbf{+64.58} \\ \hline
 \multirow{3}{*}{$K=5$}&Retrain from Scratch&0.190&+59.47 \\
 &Retrain from Naive MTL Init.&0.194&+58.87 \\
 &Ours (one-shot)&\textbf{0.168}&\textbf{+65.01} \\ \hline \hline
\end{tabular}}
\vspace{-3mm}
\caption{Ablation on the merits of one-shot nature for MTG. We use the same group categorization on Taskonomy-5, as identified by our method, for all trials. We perform two double-shot MTG methods, which are retrained from \emph{scratch} and \emph{Naive MTL initialization}, respectively. Other parameters are the same as Table \ref{tab:taskonomy2}.}
\vspace{-1mm}
\label{tab:taskonomy_retrain}
\end{table}

\vspace{-1mm}
\subsection{Generalizing-ability to Transformer Backbone \label{sec:transformer}} 

We perform the Taskonomy-5 experiment by replacing the variant of Xception with ViT-Base backbone \cite{dosovitskiy2020image}. For HOA, TAG, and MTG-Net, we use the same group categorization as those in Table \ref{tab:taskonomy2}. The results are shown in Table \ref{tab:vitbase}, illustrating that our method can generalize to transformer backbones, and consistently outperforms our counterparts given the same network backbone.

\begin{table}[t]
\centering
\resizebox{0.87\columnwidth}{!}{
\begin{tabular}{c|l|cc}
\hline
 Backbone&\makecell[c]{Methods}&Total Loss $\downarrow$& NormGain$_L$ (\%) \\ \hline \hline
 \multirow{5}{*}{ViT-Base}&Naive MTL&0.453&- \\
 &STL&0.435 &+58.72 \\
 &HOA &0.379 &+62.22\\
 &TAG &0.439 &+58.72\\
 &MTG-Net &0.403 &+47.65\\
 &Ours&\textbf{0.326}&\textbf{+68.31}\\ \hline \hline
\end{tabular}}
\vspace{-2mm}
\caption{Ablation on ViT-Base backbones of Taskonomy-5 with $K = 3$. Other parameters are the same as Table \ref{tab:taskonomy2}.}
\vspace{-2mm}
\label{tab:vitbase}
\end{table}

\vspace{-1mm}
\subsection{Flexibility with Amounts of Shared Layers \label{sec:sharing}}
\vspace{-1mm}

As shown in Fig. \ref{img:framework}, our method enables to share the backbone encoder layers across different groups, which naturally introduces a flexible design regarding further training efficiency (\ie, sharing more layers) and a better representation capability (\eg, sharing specific layers). 

We perform ablations with 0, 3, 6, 9 shared blocks on the 12-block Xception backbone of Taskonomy-5 with $K=3$. Table \ref{tab:shared and no shared} shows that sharing different amounts of backbone layers happens to perform comparable with each other, where sharing 3 blocks out of 12 slightly outperforms other counterparts. We also note that sharing 9 blocks delivers good results with a significantly improved efficiency, \ie, 58\% FLOPs \emph{w.r.t.} the model without any backbone sharing.

\begin{table}[t]
\centering
\resizebox{0.92\columnwidth}{!}{
\begin{tabular}{c|ccc}
\hline
Shared Blocks & Total Loss $\downarrow$  & $\text{NormGain}_{L}$ (\%) $\uparrow$ & FLOPs (G) \\ \hline \hline
0             & 0.174 & +63.72         & 56.89     \\
3             & \textbf{0.166} & \textbf{+65.64}         & 49.68      \\  
6             & 0.173 & +63.85         & 42.67     \\ 
9             & 0.169 & +64.75         & \textbf{33.36}    \\ \hline \hline
\end{tabular}}
\vspace{-2mm}
\caption{Ablation on different amounts of shared backbone layers. The results are obtained on Taskonomy-5 with 12-block Xception backbone and $K=3$. Other parameters are the same as Table \ref{tab:taskonomy2}.}
\label{tab:shared and no shared}
\vspace{-5mm}
\end{table}

\vspace{-2mm}
\section{Discussion and Conclusion}
\vspace{-1mm}

\textbf{Limitations and Future Works.} Our method has the following two limitations. 1) The Gumbel Softmax is biased \cite{grathwohl2018backpropagation,tucker2017rebar} and may be sensitive to different temperatures. We thus fix the temperature as a small value like \cite{chen2018learning} to alleviate the bias issue, and we empirically show in Appendix \ref{appeix:tau} that different temperatures do not alter the performance significantly for our problem. We note that Gumbel Softmax is not our contribution and we will seek alternative optimizers in the future. 2) Our training efficiency can be impeded in uncommon cases where the network heads dominate the computations of the whole network (due to the expended $KN$ heads), as discussed in Appendix \ref{appeix:time}. A more efficient training strategy for heavy heads is desirable in the future.

\textbf{Conclusion.} We tackle the challenges of Multi-Task Learning with numerous tasks using Multi-Task Grouping (MTG) techniques. Our approach efficiently identifies the best task groups from $2^N$ candidates given $N$ input tasks, with the high-order task affinity fully exploited. Moreover, our unified training approach of group identification and grouped task learning can be directly optimized using the task losses in one shot, which further improves the training efficiency and mitigates potential bias in separate training. We formulate MTG as a pruning problem, where the pruning process is also efficient as only the task heads are expended for pruning, instead of expanding the heavy encoders. We validate our methods on CelebA and Taskonomy datasets with extensive ablations. The results demonstrate the promising performance and the desirable efficiency of our method.

\section*{Acknowledgements}

This work was supported by the National Natural Science Foundation of China (62306214, 62325111, 62076099), the Natural Science Foundation of Hubei Province of China (2023AFB196), and the Knowledge Innovation Program of Wuhan-Shuguang Project (2023010201020258).

\section*{Impact Statement}

This paper presents a novel one-shot differentiable method for general-purpose multi-task grouping. There is no potential societal consequence that must be specifically highlighted here.

\bibliography{main}

\begin{thebibliography}{81}
\providecommand{\natexlab}[1]{#1}
\providecommand{\url}[1]{\texttt{#1}}
\expandafter\ifx\csname urlstyle\endcsname\relax
  \providecommand{\doi}[1]{doi: #1}\else
  \providecommand{\doi}{doi: \begingroup \urlstyle{rm}\Url}\fi

\bibitem[Akimoto et~al.(2019)Akimoto, Shirakawa, Yoshinari, Uchida, Saito, and
  Nishida]{akimotoadaptive2019}
Akimoto, Y., Shirakawa, S., Yoshinari, N., Uchida, K., Saito, S., and Nishida,
  K.
\newblock Adaptive stochastic natural gradient method for one-shot neural
  architecture search.
\newblock In \emph{ICML}, pp.\  171--180, 2019.

\bibitem[Argyriou et~al.(2008)Argyriou, Evgeniou, and Pontil]{convexmtl2008}
Argyriou, A., Evgeniou, T., and Pontil, M.
\newblock Convex multi-task feature learning.
\newblock \emph{Machine learning}, 73:\penalty0 243--272, 2008.

\bibitem[Bergstra et~al.(2013)Bergstra, Yamins, and Cox]{makingbergstra2013}
Bergstra, J., Yamins, D., and Cox, D.
\newblock Making a science of model search: Hyperparameter optimization in
  hundreds of dimensions for vision architectures.
\newblock In \emph{ICML}, pp.\  115--123, 2013.

\bibitem[Bertoni et~al.(2019)Bertoni, Kreiss, and Alahi]{bertonimonoloco2019}
Bertoni, L., Kreiss, S., and Alahi, A.
\newblock Monoloco: Monocular 3d pedestrian localization and uncertainty
  estimation.
\newblock In \emph{ICCV}, pp.\  6861--6871, 2019.

\bibitem[Bilen \& Vedaldi(2016)Bilen and Vedaldi]{mtl2016}
Bilen, H. and Vedaldi, A.
\newblock Integrated perception with recurrent multi-task neural networks.
\newblock In \emph{NeurIPS}, volume~29, 2016.

\bibitem[Br{\"u}ggemann et~al.(2020)Br{\"u}ggemann, Kanakis, Georgoulis, and
  Gool]{bruggemann2020}
Br{\"u}ggemann, D., Kanakis, M., Georgoulis, S., and Gool, L.~V.
\newblock Automated search for resource-efficient branched multi-task networks.
\newblock In \emph{BMVC}, pp.\  359, 2020.

\bibitem[Caesar et~al.(2020)Caesar, Bankiti, Lang, Vora, Liong, Xu, Krishnan,
  Pan, Baldan, and Beijbom]{caesarnuscenes2020}
Caesar, H., Bankiti, V., Lang, A., Vora, S., Liong, V., Xu, Q., Krishnan, A.,
  Pan, Y., Baldan, G., and Beijbom, O.
\newblock nuscenes: A multimodal dataset for autonomous driving.
\newblock In \emph{CVPR}, pp.\  11621--11631, 2020.

\bibitem[Cai et~al.(2018)Cai, Zhu, and Han]{caiproxylessnas2018}
Cai, H., Zhu, L., and Han, S.
\newblock Proxylessnas: Direct neural architecture search on target task and
  hardware.
\newblock In \emph{ICLR}, 2018.

\bibitem[Caruana(1997)]{mtl1997}
Caruana, R.
\newblock Multitask learning.
\newblock \emph{Machine learning}, 28:\penalty0 41--75, 1997.

\bibitem[Chen et~al.(2018{\natexlab{a}})Chen, Song, Wainwright, and
  Jordan]{chen2018learning}
Chen, J., Song, L., Wainwright, M.~J., and Jordan, M.~I.
\newblock Learning to explain: An information-theoretic perspective on model
  interpretation.
\newblock In \emph{ICML}, 2018{\natexlab{a}}.

\bibitem[Chen et~al.(2018{\natexlab{b}})Chen, Meng, Zhang, Xiang, Huang, Mu,
  and Wang]{chenreinforced2018}
Chen, Y., Meng, G., Zhang, Q., Xiang, S., Huang, C., Mu, L., and Wang, X.
\newblock Reinforced evolutionary neural architecture search.
\newblock \emph{arXiv preprint arXiv:1808.00193}, 2018{\natexlab{b}}.

\bibitem[Chen et~al.(2018{\natexlab{c}})Chen, Badrinarayanan, Lee, and
  Rabinovich]{gradnorm12018}
Chen, Z., Badrinarayanan, V., Lee, C., and Rabinovich, A.
\newblock Gradnorm: Gradient normalization for adaptive loss balancing in deep
  multitask networks.
\newblock In \emph{ICML}, pp.\  794--803, 2018{\natexlab{c}}.

\bibitem[Chen et~al.(2020)Chen, Ngiam, Huang, Luong, Kretzschmar, Chai, and
  Anguelov]{weightingtask22020}
Chen, Z., Ngiam, J., Huang, Y., Luong, T., Kretzschmar, H., Chai, Y., and
  Anguelov, D.
\newblock Just pick a sign: Optimizing deep multitask models with gradient sign
  dropout.
\newblock \emph{NeurIPS}, 33:\penalty0 2039--2050, 2020.

\bibitem[Cheng et~al.(2021)Cheng, Schwing, and Kirillov]{maskformer2021}
Cheng, B., Schwing, A., and Kirillov, A.
\newblock Per-pixel classification is not all you need for semantic
  segmentation.
\newblock \emph{NeurIPS}, 34:\penalty0 17864--17875, 2021.

\bibitem[Chollet(2017)]{xception2017}
Chollet, F.
\newblock Xception: Deep learning with depthwise separable convolutions.
\newblock In \emph{CVPR}, pp.\  1251--1258, 2017.

\bibitem[Dosovitskiy et~al.(2021)Dosovitskiy, Beyer, Kolesnikov, Weissenborn,
  Zhai, Unterthiner, Dehghani, Minderer, Heigold, Gelly,
  et~al.]{dosovitskiy2020image}
Dosovitskiy, A., Beyer, L., Kolesnikov, A., Weissenborn, D., Zhai, X.,
  Unterthiner, T., Dehghani, M., Minderer, M., Heigold, G., Gelly, S., et~al.
\newblock An image is worth 16x16 words: Transformers for image recognition at
  scale.
\newblock In \emph{ICLR}, 2021.

\bibitem[Dvornik et~al.(2017)Dvornik, Shmelkov, Mairal, and
  Schmid]{blitznet2017}
Dvornik, N., Shmelkov, K., Mairal, J., and Schmid, C.
\newblock Blitznet: A real-time deep network for scene understanding.
\newblock In \emph{ICCV}, pp.\  4154--4162, 2017.

\bibitem[Elsken et~al.(2019)Elsken, Metzen, and Hutter]{elskenneural2019}
Elsken, T., Metzen, J., and Hutter, F.
\newblock Neural architecture search: A survey.
\newblock \emph{The Journal of Machine Learning Research}, 20\penalty0
  (1):\penalty0 1997--2017, 2019.

\bibitem[Fifty et~al.(2021)Fifty, Amid, Zhao, Yu, Anil, and Finn]{tag2021}
Fifty, C., Amid, E., Zhao, Z., Yu, T., Anil, R., and Finn, C.
\newblock Efficiently identifying task groupings for multi-task learning.
\newblock In \emph{NeurIPS}, volume~34, pp.\  27503--27516, 2021.

\bibitem[Gao et~al.(2019)Gao, Ma, Zhao, Liu, and Yuille]{nddr2019}
Gao, Y., Ma, J., Zhao, M., Liu, W., and Yuille, A.
\newblock Nddr-cnn: Layerwise feature fusing in multi-task cnns by neural
  discriminative dimensionality reduction.
\newblock In \emph{CVPR}, pp.\  3205--3214, 2019.

\bibitem[Gao et~al.(2020)Gao, Bai, Jie, Ma, Jia, and Liu]{gao2020mtl}
Gao, Y., Bai, H., Jie, Z., Ma, J., Jia, K., and Liu, W.
\newblock {MTL}-{NAS}: Task-agnostic neural architecture search towards
  general-purpose multi-task learning.
\newblock In \emph{CVPR}, 2020.

\bibitem[Gao et~al.(2024)Gao, Zhang, Luo, Ma, Yu, Xia, and Ma]{gao2024aux}
Gao, Y., Zhang, W., Luo, W., Ma, L., Yu, J.-G., Xia, G.-S., and Ma, J.
\newblock {Aux}-{NAS}: Exploiting auxiliary labels with negligibly extra
  inference cost.
\newblock In \emph{ICLR}, 2024.

\bibitem[Ghiasi et~al.(2019)Ghiasi, Lin, and Le]{ghiasifpn2019}
Ghiasi, G., Lin, T., and Le, Q.
\newblock Nas-fpn: Learning scalable feature pyramid architecture for object
  detection.
\newblock In \emph{CVPR}, pp.\  7036--7045, 2019.

\bibitem[Godard et~al.(2019)Godard, Aodha, Firman, and
  Brostow]{godarddigging2019}
Godard, C., Aodha, A.~M., Firman, M., and Brostow, G.
\newblock Digging into self-supervised monocular depth estimation.
\newblock In \emph{CVPR}, pp.\  3828--3838, 2019.

\bibitem[Gordon et~al.(2018)Gordon, Eban, Nachum, Chen, Wu, Yang, and
  Choi]{gordonmorphnet2018}
Gordon, A., Eban, E., Nachum, O., Chen, B., Wu, H., Yang, T., and Choi, E.
\newblock Morphnet: Fast \& simple resource-constrained structure learning of
  deep networks.
\newblock In \emph{CVPR}, pp.\  1586--1595, 2018.

\bibitem[Grathwohl et~al.(2018)Grathwohl, Choi, Wu, Roeder, and
  Duvenaud]{grathwohl2018backpropagation}
Grathwohl, W., Choi, D., Wu, Y., Roeder, G., and Duvenaud, D.
\newblock Backpropagation through the void: Optimizing control variates for
  black-box gradient estimation.
\newblock In \emph{ICLR}, 2018.

\bibitem[Guo et~al.(2018)Guo, Haque, Huang, Yeung, and Li]{rearranging2018}
Guo, M., Haque, A., Huang, D., Yeung, S., and Li, F.
\newblock Dynamic task prioritization for multitask learning.
\newblock In \emph{ECCV}, pp.\  270--287, 2018.

\bibitem[Guo et~al.(2019)Guo, Zhong, Wu, Lin, and Yan]{irlasguo2019}
Guo, M., Zhong, Z., Wu, W., Lin, D., and Yan, J.
\newblock Irlas: Inverse reinforcement learning for architecture search.
\newblock In \emph{CVPR}, pp.\  9021--9029, 2019.

\bibitem[Guo et~al.(2020)Guo, Lee, and Ulbricht]{learntobranch2020}
Guo, P., Lee, C., and Ulbricht, D.
\newblock Learning to branch for multi-task learning.
\newblock In \emph{ICML}, pp.\  3854--3863. PMLR, 2020.

\bibitem[He et~al.(2020)He, Ye, Shen, and Zhang]{hemilenas2020}
He, C., Ye, H., Shen, L., and Zhang, T.
\newblock Milenas: Efficient neural architecture search via mixed-level
  reformulation.
\newblock In \emph{CVPR}, pp.\  11993--12002, 2020.

\bibitem[He et~al.(2016)He, Zhang, Ren, and Sun]{resnet2016}
He, K., Zhang, X., Ren, S., and Sun, J.
\newblock Deep residual learning for image recognition.
\newblock In \emph{CVPR}, pp.\  770--778, 2016.

\bibitem[He et~al.(2017)He, Gkioxari, Doll{\'a}r, and Girshick]{maskrcnn}
He, K., Gkioxari, G., Doll{\'a}r, P., and Girshick, R.
\newblock Mask r-cnn.
\newblock In \emph{Proceedings of the IEEE international conference on computer
  vision}, pp.\  2961--2969, 2017.

\bibitem[Hu et~al.(2023)Hu, Yang, Chen, Li, Sima, Zhu, Chai, Du, Lin, Wang,
  et~al.]{huplanning2023}
Hu, Y., Yang, J., Chen, L., Li, K., Sima, C., Zhu, X., Chai, S., Du, S., Lin,
  T., Wang, W., et~al.
\newblock Planning-oriented autonomous driving.
\newblock In \emph{CVPR}, pp.\  17853--17862, 2023.

\bibitem[Kang et~al.(2011)Kang, Grauman, and Sha]{mtl2011}
Kang, Z., Grauman, K., and Sha, F.
\newblock Learning with whom to share in multi-task feature learning.
\newblock In \emph{ICML}, pp.\  521--528, 2011.

\bibitem[Kendall et~al.(2018)Kendall, Gal, and Cipolla]{uncertainty2018}
Kendall, A., Gal, Y., and Cipolla, R.
\newblock Multi-task learning using uncertainty to weigh losses for scene
  geometry and semantics.
\newblock In \emph{CVPR}, pp.\  7482--7491, 2018.

\bibitem[Kokkinos(2017)]{ubernet2017}
Kokkinos, I.
\newblock Ubernet: Training a universal convolutional neural network for low-,
  mid-, and high-level vision using diverse datasets and limited memory.
\newblock In \emph{CVPR}, pp.\  6129--6138, 2017.

\bibitem[Kumar \& III(2012)Kumar and III]{learningtg2012}
Kumar, A. and III, H.~D.
\newblock Learning task grouping and overlap in multi-task learning.
\newblock In \emph{ICML}, pp.\  1723--1730, 2012.

\bibitem[Kurin et~al.(2022)Kurin, Palma, Kostrikov, Whiteson, and
  Mudigonda]{kurin2022}
Kurin, V., Palma, A.~D., Kostrikov, I., Whiteson, S., and Mudigonda, P.
\newblock In defense of the unitary scalarization for deep multi-task learning.
\newblock \emph{NeurIPS}, 35:\penalty0 12169--12183, 2022.

\bibitem[Li et~al.(2021)Li, Gao, and Sang]{jointgroups2021}
Li, M., Gao, Y., and Sang, N.
\newblock Exploiting learnable joint groups for hand pose estimation.
\newblock In \emph{AAAI}, volume~35, pp.\  1921--1929, 2021.

\bibitem[Li \& Gong(2021)Li and Gong]{lirobust2021}
Li, X. and Gong, H.
\newblock Robust optimization for multilingual translation with imbalanced
  data.
\newblock \emph{NeurIPS}, 34:\penalty0 25086--25099, 2021.

\bibitem[Li et~al.(2019)Li, Zhou, Pan, and Feng]{lipartial2019}
Li, X., Zhou, Y., Pan, Z., and Feng, J.
\newblock Partial order pruning: for best speed/accuracy trade-off in neural
  architecture search.
\newblock In \emph{CVPR}, pp.\  9145--9153, 2019.

\bibitem[Lin et~al.(2022)Lin, Ye, Zhang, and Tsang]{weightingtask12022}
Lin, B., Ye, F., Zhang, Y., and Tsang, I.
\newblock Reasonable effectiveness of random weighting: A litmus test for
  multi-task learning.
\newblock \emph{Transactions on Machine Learning Research}, 2022.

\bibitem[Lin et~al.(2019)Lin, Zhen, Li, Zhang, and Kwong]{lin2019}
Lin, X., Zhen, H., Li, Z., Zhang, Q., and Kwong, S.
\newblock Pareto multi-task learning.
\newblock \emph{NeurIPS}, 32, 2019.

\bibitem[Liu et~al.(2021)Liu, Liu, Jin, Stone, and Liu]{liuconflict2021}
Liu, B., Liu, X., Jin, X., Stone, P., and Liu, Q.
\newblock Conflict-averse gradient descent for multi-task learning.
\newblock \emph{NeurIPS}, 34:\penalty0 18878--18890, 2021.

\bibitem[Liu et~al.(2019{\natexlab{a}})Liu, Chen, Schroff, Adam, Hua, Yuille,
  and Li]{liuauto2019}
Liu, C., Chen, L., Schroff, F., Adam, H., Hua, W., Yuille, A., and Li, F.
\newblock Auto-deeplab: Hierarchical neural architecture search for semantic
  image segmentation.
\newblock In \emph{CVPR}, pp.\  82--92, 2019{\natexlab{a}}.

\bibitem[Liu et~al.(2020)Liu, Li, Kuang, Xue, Chen, Yang, Liao, and
  Zhang]{liutowards2020}
Liu, L., Li, Y., Kuang, Z., Xue, J., Chen, Y., Yang, W., Liao, Q., and Zhang,
  W.
\newblock Towards impartial multi-task learning.
\newblock In \emph{ICLR}, 2020.

\bibitem[Liu et~al.(2019{\natexlab{b}})Liu, Johns, and Davison]{gradnorm22019}
Liu, S., Johns, E., and Davison, A.
\newblock End-to-end multi-task learning with attention.
\newblock In \emph{CVPR}, pp.\  1871--1880, 2019{\natexlab{b}}.

\bibitem[Liu et~al.(2015)Liu, Luo, Wang, and Tang]{faceattribute2015}
Liu, Z., Luo, P., Wang, X., and Tang, X.
\newblock Deep learning face attributes in the wild.
\newblock In \emph{ICCV}, pp.\  3730--3738, 2015.

\bibitem[Long et~al.(2015)Long, Cohn, Bird, and Cook]{crosslingual2015}
Long, D., Cohn, T., Bird, S., and Cook, P.
\newblock Low resource dependency parsing: Cross-lingual parameter sharing in a
  neural network parser.
\newblock In \emph{ACL}, pp.\  845--850, 2015.

\bibitem[Long et~al.(2017)Long, Cao, Wang, and Yu]{mrn2017}
Long, M., Cao, Z., Wang, J., and Yu, P.
\newblock Learning multiple tasks with multilinear relationship networks.
\newblock In \emph{NeurIPS}, volume~30, 2017.

\bibitem[Maddison et~al.(2016)Maddison, Mnih, and Teh]{maddison2016concrete}
Maddison, C.~J., Mnih, A., and Teh, Y.~W.
\newblock The concrete distribution: A continuous relaxation of discrete random
  variables.
\newblock \emph{arXiv preprint arXiv:1611.00712}, 2016.

\bibitem[Maninis et~al.(2019)Maninis, Radosavovic, and
  Kokkinos]{maninis2019attentive}
Maninis, K.-K., Radosavovic, I., and Kokkinos, I.
\newblock Attentive single-tasking of multiple tasks.
\newblock In \emph{Proceedings of the IEEE/CVF conference on computer vision
  and pattern recognition}, pp.\  1851--1860, 2019.

\bibitem[Misra et~al.(2016)Misra, Shrivastava, Gupta, and
  Hebert]{crossstitch2016}
Misra, I., Shrivastava, A., Gupta, A., and Hebert, M.
\newblock Cross-stitch networks for multi-task learning.
\newblock In \emph{CVPR}, pp.\  3994--4003, 2016.

\bibitem[Nekrasov et~al.(2019)Nekrasov, Dharmasiri, Spek, Drummond, Shen, and
  Reid]{lwr2019}
Nekrasov, V., Dharmasiri, T., Spek, A., Drummond, T., Shen, C., and Reid, I.
\newblock Real-time joint semantic segmentation and depth estimation using
  asymmetric annotations.
\newblock In \emph{ICRA}, pp.\  7101--7107, 2019.

\bibitem[Real et~al.(2019)Real, Aggarwal, Huang, and Le]{regularizedreal2019}
Real, E., Aggarwal, A., Huang, Y., and Le, Q.
\newblock Regularized evolution for image classifier architecture search.
\newblock In \emph{AAAI}, volume~33, pp.\  4780--4789, 2019.

\bibitem[Ruder(2017)]{mtaoverview2017}
Ruder, S.
\newblock An overview of multi-task learning in deep neural networks.
\newblock \emph{arXiv preprint arXiv:1706.05098}, 2017.

\bibitem[Ruder et~al.(2019)Ruder, Bingel, Augenstein, and
  S{\o}gaard]{ruderlatent2019}
Ruder, S., Bingel, J., Augenstein, I., and S{\o}gaard, A.
\newblock Latent multi-task architecture learning.
\newblock In \emph{AAAI}, volume~33, pp.\  4822--4829, 2019.

\bibitem[Sener \& Koltun(2018)Sener and Koltun]{sener2018}
Sener, O. and Koltun, V.
\newblock Multi-task learning as multi-objective optimization.
\newblock \emph{NeurIPS}, 31, 2018.

\bibitem[Song et~al.(2022)Song, Zheng, Cao, Yu, and Bian]{mtg2022}
Song, X., Zheng, S., Cao, W., Yu, J., and Bian, J.
\newblock Efficient and effective multi-task grouping via meta learning on task
  combinations.
\newblock In \emph{NeurIPS}, volume~35, pp.\  37647--37659, 2022.

\bibitem[Standley et~al.(2020)Standley, Zamir, Chen, Guibas, Malik, and
  Savarese]{hoa2020}
Standley, T., Zamir, A., Chen, D., Guibas, L., Malik, J., and Savarese, S.
\newblock Which tasks should be learned together in multi-task learning?
\newblock In \emph{ICML}, pp.\  9120--9132, 2020.

\bibitem[Sun et~al.(2020)Sun, Panda, Feris, and k.~Saenko]{adashare2020}
Sun, X., Panda, R., Feris, R., and k.~Saenko.
\newblock Adashare: Learning what to share for efficient deep multi-task
  learning.
\newblock In \emph{NeurIPS}, volume~33, pp.\  8728--8740, 2020.

\bibitem[Suteu \& Guo(2019)Suteu and Guo]{suteuregularizing2019}
Suteu, M. and Guo, Y.
\newblock Regularizing deep multi-task networks using orthogonal gradients.
\newblock \emph{arXiv preprint arXiv:1912.06844}, 2019.

\bibitem[Tan et~al.(2019)Tan, Chen, Pang, Vasudevan, Sandler, Howard, and
  Le]{mnasnettan2019}
Tan, M., Chen, B., Pang, R., Vasudevan, V., Sandler, M., Howard, A., and Le, Q.
\newblock Mnasnet: Platform-aware neural architecture search for mobile.
\newblock In \emph{CVPR}, pp.\  2820--2828, 2019.

\bibitem[Tang et~al.(2021)Tang, Li, and Liu]{tangreview2021}
Tang, J., Li, S., and Liu, P.
\newblock A review of lane detection methods based on deep learning.
\newblock \emph{Pattern Recognition}, 111:\penalty0 107623, 2021.

\bibitem[Tucker et~al.(2017)Tucker, Mnih, Maddison, Lawson, and
  Sohl-Dickstein]{tucker2017rebar}
Tucker, G., Mnih, A., Maddison, C.~J., Lawson, D., and Sohl-Dickstein, J.
\newblock {REBAR}: Low-variance, unbiased gradient estimates for discrete
  latent variable models.
\newblock In \emph{NIPS}, 2017.

\bibitem[Vandenhende et~al.(2019)Vandenhende, Georgoulis, Brabandere, and
  Gool]{vandenhende2019}
Vandenhende, S., Georgoulis, S., Brabandere, B.~D., and Gool, L.~V.
\newblock Branched multi-task networks: Deciding what layers to share.
\newblock \emph{BMVC}, 2019.

\bibitem[Vandenhende et~al.(2020)Vandenhende, Georgoulis, and
  Van~Gool]{vandenhende2020mti}
Vandenhende, S., Georgoulis, S., and Van~Gool, L.
\newblock Mti-net: Multi-scale task interaction networks for multi-task
  learning.
\newblock In \emph{Computer Vision--ECCV 2020: 16th European Conference,
  Glasgow, UK, August 23--28, 2020, Proceedings, Part IV 16}, pp.\  527--543.
  Springer, 2020.

\bibitem[Vandenhende et~al.(2021)Vandenhende, Georgoulis, Gansbeke, Proesmans,
  Dai, and Gool]{mtaoverreview2021}
Vandenhende, S., Georgoulis, S., Gansbeke, W.~V., Proesmans, M., Dai, D., and
  Gool, L.~V.
\newblock Multi-task learning for dense prediction tasks: A survey.
\newblock \emph{IEEE TPAMI}, 44\penalty0 (7):\penalty0 3614--3633, 2021.

\bibitem[Wang \& Tsvetkov(2021)Wang and Tsvetkov]{wanggradient2021}
Wang, Z. and Tsvetkov, Y.
\newblock Gradient vaccine: Investigating and improving multi-task optimization
  in massively multilingual models.
\newblock In \emph{ICLR}, 2021.

\bibitem[Wu et~al.(2019)Wu, Dai, Zhang, Wang, Sun, Wu, Tian, Vajda, Jia, and
  Keutzer]{wufbnet2019}
Wu, B., Dai, X., Zhang, P., Wang, Y., Sun, F., Wu, Y., Tian, Y., Vajda, P.,
  Jia, Y., and Keutzer, K.
\newblock Fbnet: Hardware-aware efficient convnet design via differentiable
  neural architecture search.
\newblock In \emph{CVPR}, pp.\  10734--10742, 2019.

\bibitem[Xie \& Yuille(2017)Xie and Yuille]{xiegenetic2017}
Xie, L. and Yuille, A.
\newblock Genetic cnn.
\newblock In \emph{ICCV}, pp.\  1379--1388, 2017.

\bibitem[Xin et~al.(2022)Xin, Ghorbani, Gilmer, Garg, and Firat]{xin2022}
Xin, D., Ghorbani, B., Gilmer, J., Garg, A., and Firat, O.
\newblock Do current multi-task optimization methods in deep learning even
  help?
\newblock \emph{NeurIPS}, 35:\penalty0 13597--13609, 2022.

\bibitem[Xu et~al.(2018)Xu, Ouyang, Wang, and Sebe]{padnet2018}
Xu, D., Ouyang, W., Wang, X., and Sebe, N.
\newblock Pad-net: Multi-tasks guided prediction-and-distillation network for
  simultaneous depth estimation and scene parsing.
\newblock In \emph{CVPR}, pp.\  675--684, 2018.

\bibitem[Yang \& Hospedales(2016)Yang and Hospedales]{yangdeep2016}
Yang, Y. and Hospedales, T.
\newblock Deep multi-task representation learning: A tensor factorisation
  approach.
\newblock In \emph{ICLR}, 2016.

\bibitem[Yu et~al.(2020)Yu, Kumar, Gupta, Levine, Hausman, and
  Finn]{yugradient2020}
Yu, T., Kumar, S., Gupta, A., Levine, S., Hausman, K., and Finn, C.
\newblock Gradient surgery for multi-task learning.
\newblock \emph{NeurIPS}, 33:\penalty0 5824--5836, 2020.

\bibitem[Zamir et~al.(2016)Zamir, Wekel, Agrawal, Wei, Malik, and
  Savarese]{mtl3d2016}
Zamir, A., Wekel, T., Agrawal, P., Wei, C., Malik, J., and Savarese, S.
\newblock Generic 3d representation via pose estimation and matching.
\newblock In \emph{ECCV}, pp.\  535--553. Springer, 2016.

\bibitem[Zamir et~al.(2018)Zamir, Sax, Shen, Guibas, Malik, and
  Savarese]{taskonomy2018}
Zamir, A., Sax, A., Shen, W., Guibas, L., Malik, J., and Savarese, S.
\newblock Taskonomy: Disentangling task transfer learning.
\newblock In \emph{CVPR}, pp.\  3712--3722, 2018.

\bibitem[Zhang et~al.(2019{\natexlab{a}})Zhang, Qiu, Liu, Yao, Liu, and
  Mei]{zhangcustomizable2019}
Zhang, Y., Qiu, Z., Liu, J., Yao, T., Liu, D., and Mei, T.
\newblock Customizable architecture search for semantic segmentation.
\newblock In \emph{CVPR}, pp.\  11641--11650, 2019{\natexlab{a}}.

\bibitem[Zhang et~al.(2018)Zhang, Cui, Xu, Jie, Li, and
  J.~Yang]{zhangjoint2018}
Zhang, Z., Cui, Z., Xu, C., Jie, Z., Li, X., and J.~Yang, J.
\newblock Joint task-recursive learning for semantic segmentation and depth
  estimation.
\newblock In \emph{ECCV}, pp.\  235--251, 2018.

\bibitem[Zhang et~al.(2019{\natexlab{b}})Zhang, Cui, Xu, Yan, Sebe, and
  Yang]{zhangpattern2019}
Zhang, Z., Cui, Z., Xu, C., Yan, Y., Sebe, N., and Yang, J.
\newblock Pattern-affinitive propagation across depth, surface normal and
  semantic segmentation.
\newblock In \emph{CVPR}, pp.\  4106--4115, 2019{\natexlab{b}}.

\bibitem[Zoph et~al.(2018)Zoph, Vasudevan, Shlens, and Le]{learningzoph2018}
Zoph, B., Vasudevan, V., Shlens, J., and Le, Q.
\newblock Learning transferable architectures for scalable image recognition.
\newblock In \emph{CVPR}, pp.\  8697--8710, 2018.

\end{thebibliography}
\bibliographystyle{icml2024}

\newpage
\appendix
\onecolumn

\setcounter{table}{0}
\setcounter{figure}{0}
\setcounter{equation}{0}

\renewcommand{\theequation}{A\arabic{equation}}
\renewcommand{\thefigure}{A\arabic{figure}}
\renewcommand{\thetable}{A\arabic{table}}

\section{Datasets \label{appdix:dataset}}
\begin{figure}[h]
  \centering
  \subfigure[Illustration samples of the CelebA datasets\label{img:celeba}]{\includegraphics[width=0.37\columnwidth]{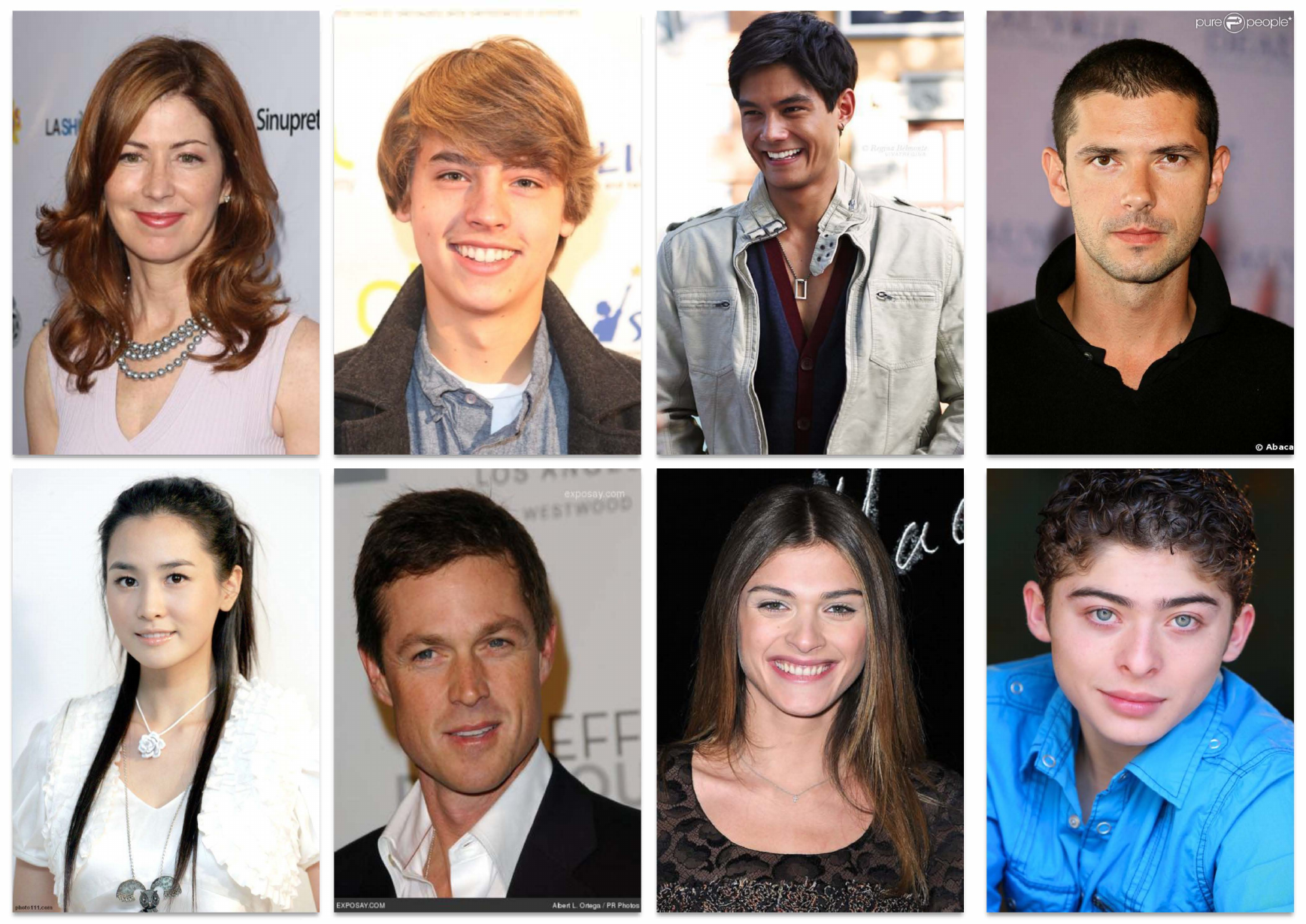}}
  \hspace{0.3cm} 
  \subfigure[Illustration samples of the Taskonomy datasets\label{img:taskonomy}]{\includegraphics[width=0.37\columnwidth, height=4.46cm]{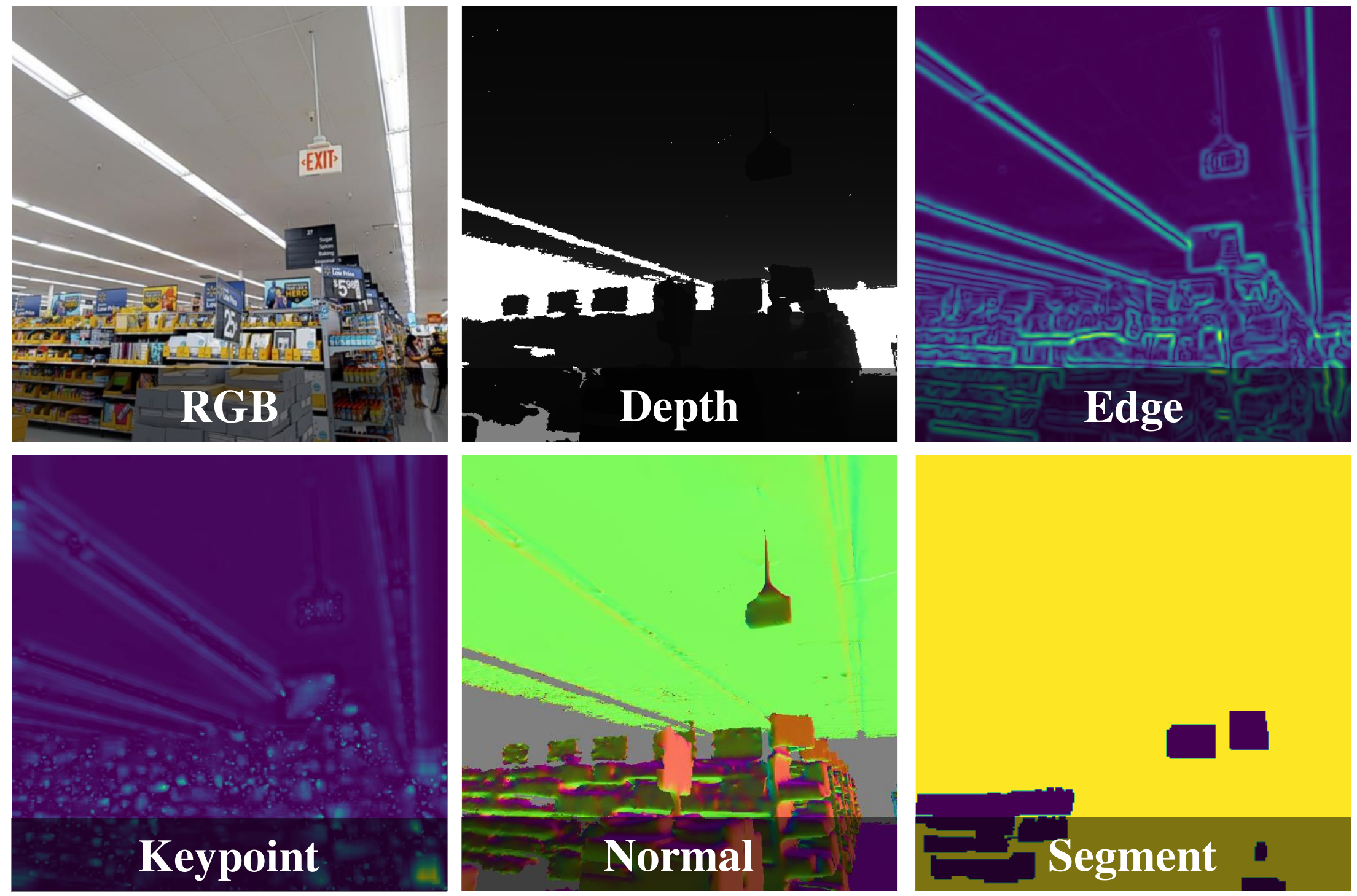}}
  \caption{Illustration of the CelebA and Taskonomy datasets. Several image samples across different genders and races of CelebA dataset are shown in subfigure \subref{img:celeba}, while subfigure \subref{img:taskonomy} exhibits an indoor RGB image with 5 annotated labels used in our experiments.}
\end{figure}

\noindent \textbf{CelebA} \cite{faceattribute2015} is a large-scale face dataset that contains more than $200,000$ images from roughly $10,000$ identities, each of which has annotated $40$ face attributes representing the tasks to predict. Samples of CelebA datasets are illustrated in Fig. \ref{img:celeba}. For the CelebA experiments, the $9$ tasks implemented in CelebA-$9$ \cite{tag2021} are \textit{5\_o\_Clock\_Shadow}, \textit{Black\_Hair}, \textit{Blond\_Hair}, \textit{Brown\_Hair}, \textit{Goatee}, \textit{Mustache}, \textit{No\_Beard}, \textit{Rosy\_Cheeks}, and \textit{Wearing\_Hat}. We also test our method with CelebA-$40$ using all the 40 labeled attributes of the CelebA dataset. All images in the CelebA dataset are resized to $64\times64$ resolution.

\textbf{Taskonomy} \cite{taskonomy2018} is one of the largest datasets with multi-task labels, covering 26 vision tasks from 2D to 3D. For the Taskonomy experiments, we use the official tiny train, validation, and test split with roughly $300,000$ images. The tasks used to learn the categorization are the same as those in \cite{hoa2020,tag2021,mtg2022}, which are \textit{semantic segmentation}, \textit{depth estimation}, \textit{surface normal}, \textit{keypoint detection}, and \textit{canny edge detection}, as shown in Figure \ref{img:taskonomy}. We also follow \cite{hoa2020,tag2021,mtg2022} to bilinearly downsample the Taskonomy images to $256\times 256$ before training and testing. 




\section{The Training Time, FLOPs, and the Influence of Heavy Heads \label{appeix:time}}

We illustrate the training time of all the methods for different experiments and network backbones in Table \ref{tab:time}. Table \ref{tab:time} shows that our method outperforms the previous arts in terms of both FLOPs and training time. We also note that for uncommon cases where the network heads dominate the computations of the whole network, \ie, Taskonomy-9 experiments with the Xception Variant backbone, the efficiency of our method will be impeded by those heavy heads, as we expand $N$ heads to $KN$ (given $N$ tasks and $K$ groups). We leave the efficiency training strategy for heavy heads as our future work.

\begin{table}[h]
\centering
\resizebox{\columnwidth}{!}{
\begin{tabular}{l|l|c|c||c|c|c|c|c|c}
\hline \hline
\multirow{2}{*}{Experiment} & \multirow{2}{*}{Network Backbone} & \multirow{2}{*}{Heads FLOPs Portion} & \multirow{2}{*}{Groups} & \multicolumn{2}{c|}{HOA}               & \multicolumn{2}{c|}{TAG}            & \multicolumn{2}{c}{Ours}                                       \\
\cline{5-10}
           &                  &                        &        & FLOPs (G) & Time (h) & FLOPs (G) & Time (h) & FLOPs (G)              & Time (h)               \\
\hline \hline
\multirow{3}{*}{CelebA-9}   & \multirow{3}{*}{ResNet Variant}   & \multirow{3}{*}{0.32\%}                 & $K=2$    & 3.02      & 18.42    & 0.24      & 4.79     & \textbf{0.15}          & \textbf{0.99}          \\
           &                  &                        & $K=3$    & 3.10      & 18.90    & 0.32      & 5.27     & \textbf{\textbf{0.22}} & \textbf{\textbf{1.27}} \\
           &                  &                        & $K=4$    & 3.18      & 19.38    & 0.40      & 5.70     & \textbf{\textbf{0.29}} & \textbf{\textbf{1.55}} \\
\hline
\multirow{3}{*}{Taskonomy-5}  & \multirow{3}{*}{Xception Variant} & \multirow{3}{*}{59.73\%}                & $K=3$    & 135.36    & 523.67   & 46.19     & 297.33   & \textbf{38.90}         & \textbf{158.56}        \\
           &                  &                        & $K=4$    & 143.96    & 555.33   & 58.16     & 345.00   & \textbf{50.68}         & \textbf{243.50}        \\
           &                  &                        & $K=5$    & 152.24    & 588.67   & 71.81     & 398.67   & \textbf{62.47}         & \textbf{328.67}   \\    
\hline
\multirow{3}{*}{Taskonomy-5}  & \multirow{3}{*}{ViT-Base} & \multirow{3}{*}{28.44\%}&$K=3$&673.99&770.00&216.51&460.33&\textbf{141.00}&\textbf{220.00} \\
    &&&$K=4$&722.12&826.00&271.53&522.33&\textbf{180.61}&\textbf{337.86} \\
    &&&$K=5$&769.76&874.81&329.93&602.16&\textbf{220.23}&\textbf{525.41} \\ 
\hline \hline
\end{tabular}}
\vspace{-2mm}
\caption{Statistics of the training time and FLOPs for different experiments and network backbones. We illustrate the GFLOPs, and the training time (hour) is obtained on a single NVIDIA 4090 GPU. Xception Variant backbone is the same as that used in HOA \cite{hoa2020} and TAG \cite{tag2021}, ResNet Variant backbone is the same as that used in TAG \cite{tag2021}. \textbf{Heads FLOPs Portion} is calculated by (the FLOPs of heads) / (the FLOPs of the whole network) of the \textbf{Naive MTL} architecture (\ie, a fully-shared encoder with $N$ task heads).}
\label{tab:time}
\end{table}

\section{Results in Terms of Errors on the Taskonomy-5 Experiments \label{appdix:error_taskonomy}}

We used the loss metric in the main text (\ie, Table \ref{tab:taskonomy_per_task}) following the Taskonomy-5 experiments in the TAG paper \cite{tag2021}. In this section, we show the error statistics of the Taskonomy-5 experiments.

Specifically, we use mean Intersection over Union (\textbf{mIoU}) for semantic segmentation, Root Mean Square Error (\textbf{RMSE}) after aligning the transformation and scale for depth estimation, the \textbf{percent of vectors with an angle less than 30 degrees for surface normal}, \textbf{F1-score} for keypoint detection, and \textbf{F1-score} for edge detection. The results are shown in Table \ref{tab:taskonomy_per_task_error}, demonstrating that our method remains the best for most tasks over the prior arts.

\begin{table*}[h]
\centering
\resizebox{0.99\textwidth}{!}{
    \begin{tabular}{c|l||cc|cc|cc|cc|cc|c}
    \hline
     \makecell[c]{\multirow{2}{*}{Groups}}&\makecell[c]{\multirow{2}{*}{Methods}}& \multicolumn{2}{c|}{Depth Estimation} & \multicolumn{2}{c|}{Surface Normal} & \multicolumn{2}{c|}{Semantic Segmentation} & \multicolumn{2}{c|}{Keypoint Detection} & \multicolumn{2}{c|}{Edge Detection}&\multirow{2}{*}{Avg. NormGain$_E$}\\ \cline{3-12}
     &&RSME $\downarrow$&NormGain$_E$ (\%) $\uparrow$&Acc. ($<30^{\circ}$) $\uparrow$&NormGain$_E$ (\%) $\uparrow$&mIoU $\uparrow$&NormGain$_E$ (\%) $\uparrow$&F1 $\uparrow$&NormGain$_E$ (\%) $\uparrow$&F1 $\uparrow$&NormGain$_E$ (\%) $\uparrow$ \\ \hline
        - & Naive MTL & 8.67E-03 & - & 84.32 & - & 48.21 & - & 76.08 & - & 76.56 & - & - \\
        - & STL & 1.59E-05 & 99.82 & 84.28 & -0.05 & 37.88 & -21.44 & 85.86 & 12.86 & 86.75 & 13.31 &20.90  \\ \hline
        \multirow{6}{*}{K=3} & RG & 2.57E-02 & -195.87 & 84.35 & 0.03 & 42.84 & -11.13 & 78.71 & 3.47 & 79.71 & 4.11 &-39.88  \\
        & HOA & 5.85E-03 & 32.48 & 83.56 & -0.90 & 47.25 & -1.99 & \textbf{86.86} & \textbf{14.17} & 87.07 & 13.73 &11.50 \\
        & TAG & 5.15E-03 & 40.59 & 82.00 & -2.76 & 42.84 & -11.13 & \textbf{86.86} & \textbf{14.17} & 87.07 & 13.73 &10.92 \\
        & MTG-Net & 2.04E-04 & 97.65 & 84.32 & 0.00 & 48.21 & 0.00 & 84.47 & 11.04 & 85.98 & 12.30 &24.20 \\
        & Ours & \textbf{1.19E-07} & \textbf{100.00} & \textbf{84.52} & \textbf{0.24} & \textbf{48.93} & \textbf{1.50} & 86.34 & 13.50 & \textbf{87.94} & \textbf{14.86} & \textbf{26.02} \\ \hline
        \multirow{6}{*}{K=4} & RG & 5.15E-03 & 40.59 & 84.35 & 0.03 & 48.21 & 0.00 & 76.08 & 0.00 & 79.71 & 4.11 &8.95 \\
        & HOA & 5.15E-03 & 40.59 & 84.39 & 0.08 & 48.46 & 0.51 & \textbf{86.86} & \textbf{14.17} & 87.07 & 13.73 &13.82 \\
        & TAG & 5.15E-03 & 40.59 & 83.56 & -0.90 & 47.25 & -1.99 & \textbf{86.86} & \textbf{14.17} & 87.07 & 13.73 &13.12 \\
        & MTG-Net & 2.04E-04 & 97.65 & 84.32 & 0.00 & 48.21 & 0.00 & 84.47 & 11.04 & 85.98 & 12.30 &24.20 \\
        & Ours & \textbf{1.19E-07} & \textbf{100.00} & \textbf{84.71} & \textbf{0.46} & \textbf{49.69} & \textbf{3.08} & 86.31 & 13.46 & \textbf{87.65} & \textbf{14.49} &\textbf{26.30} \\ \hline
        \multirow{6}{*}{K=5} & RG & 5.15E-03 & 40.59 & 84.35 & 0.03 & 48.21 & 0.00 & 78.71 & 3.47 & 79.71 & 4.11 &9.64  \\
        & HOA & 5.15E-03 & 40.59 & 84.39 & 0.08 & 48.46 & 0.51 & 86.86 & 14.17 & 87.07 & 13.73 &13.82 \\
        & TAG & 5.15E-03 & 40.59 & 83.56 & -0.90 & 47.25 & -1.99 & 86.86 & 14.17 & 87.07 & 13.73 &13.12 \\
        & MTG-Net & 2.04E-04 & 97.65 & 84.32 & 0.00 & 48.21 & 0.00 & 84.47 & 11.04 & 85.98 & 12.30 &24.20 \\
        & Ours & \textbf{1.19E-07} & \textbf{100.00} & \textbf{84.40} & \textbf{0.09} & \textbf{49.85} & \textbf{3.40} & \textbf{86.90} & \textbf{14.23} & \textbf{87.94} & \textbf{14.86} &\textbf{26.52}  \\ \hline
    \end{tabular}}
    \caption{Performance (error) on Taskonomy-5 \emph{w.r.t.} each input task. Other parameters are the same as those in Table \ref{tab:taskonomy2}.}
    \label{tab:taskonomy_per_task_error}
\end{table*}

\section{Scalability to More Input Tasks \label{appdix:scalability}} 

The training complexity for the encoder of our method is $O(K)$, indicating that our training complexity is only relevant to the group number $K$ rather than the task number $N$. This is distinct from the previous state-of-the-art methods \cite{hoa2020,tag2021,mtg2022}, and therefore naturally scale to an arbitrary number of input tasks\footnote{Our complexity would be relevant to $N$ only for uncommon cases where the network heads dominate the computation of the whole network, due to the expanded $KN$ heads as discussed in Appendix \ref{appeix:time}.-*
The complexities of the previous state-of-the-art methods are all determined by $N$, as shown in Table \ref{tab:taskonomy2} and discussed in Footnote \ref{fn:MTG-Net}.}.

We validate our method with a more challenging case of categorizing all the \textbf{40} tasks into 5 groups on the CelebA dataset, denoted as CelebA-40. For the same candidate groups $K=5$, our method for CelebA-9 and CelebA-40 takes 1.7 and 2.9 GPU hours, respectively, on a single Nvidia 4090 GPU, demonstrating the scalability of our method \emph{w.r.t.} more input tasks.

\begin{figure*}[h]
    \centering
    \includegraphics[width=\textwidth]{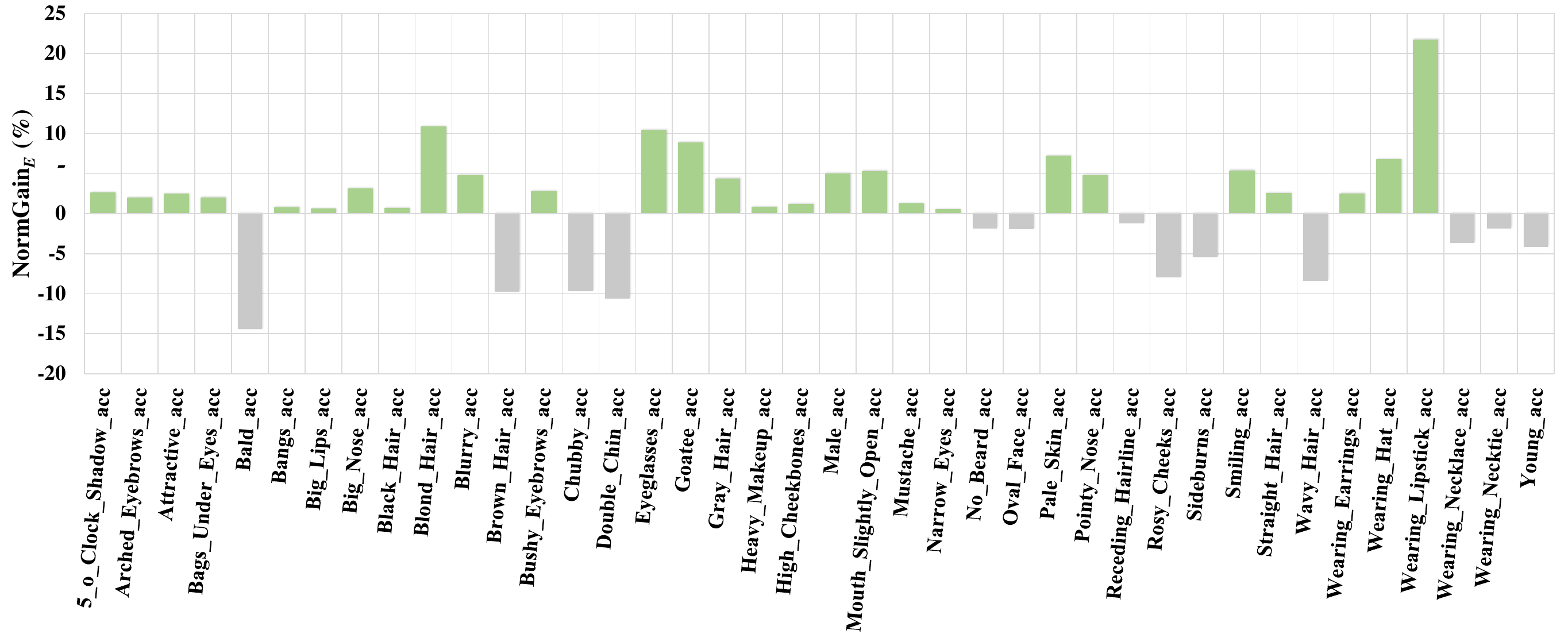}
    \caption{Normalized gain \emph{w.r.t.} classification errors, NormGain$_E$, of our method in terms of Naive MTL for each task on the CelebA dataset with $N=40$ tasks and $K=5$ groups. Other parameters are identical to those in Table \ref{tab:celeba}.}
    \label{img:celeba-40}
\end{figure*}

We illustrate the relative gain \emph{w.r.t.} Naive MTL method for each task in Fig. \ref{img:celeba-40}. Compared with Table \ref{tab:celeba}, the improvement of our method for CelebA-40 in Fig. \ref{img:celeba-40} is not as significant as that for CelebA-9 in Table \ref{tab:celeba}. This is attributed to the exponential $2^N$ difficulty in modeling the high-order task affinity, given a drastic increase of $N$ from 9 to 40. Despite that prior state-of-the-art methods suffer significant difficulties in training complexity when scaling to such an extreme case of CelebA-40, our method successfully improves most tasks for CelebA-40 by categorizing and optimizing collaborative tasks within the same group. 

\section{Influence of Gumbel Softmax Temperatures \label{appeix:tau}}

Gumbel Softmax \cite{maddison2016concrete} is known to be biased \cite{grathwohl2018backpropagation,tucker2017rebar} and can potentially be sensitive to different temperatures. In this section, we investigate different temperatures in Table \ref{tab:temp}, which demonstrates that, empirically, different temperatures do not alter the performance significantly for our problem.

\begin{table}[h]
\centering
\small
\begin{tabular}{l|c}
\hline \hline
Temperature Strategies of Gumbel Softmax             & Total Loss \\
\hline \hline
Annealing from 100 to 4 with a decay factor of 1/2   & 0.1656     \\
Annealing from 100 to 4 with a decay factor of 1/4   & 0.1665     \\
Annealing from 50 to 4 with a decay factor of 1/2    & 0.1730     \\
Annealing from 10 to 0.01 with a decay factor of 1/2 & 0.1710     \\
Annealing from 10 to 0.01 with a decay factor of 1/2 & 0.1710     \\
A fixed temperature of 4                             & 0.1684     \\
\hline \hline
\end{tabular}
\caption{Ablation on different Gumbel Softmax temperatures. The experiments are performed on Taskonomy-5 with the Xception variant backbone and $K=5$. Other parameters are identical to those in Table \ref{tab:taskonomy2}.}
\label{tab:temp}
\end{table}

\section{Discussion on Group Categorization \label{appdix:categorization}} 

We illustrate the group categorization identified by different MTG methods in Table \ref{tab:taskonomy_grouping} and Table \ref{tab:celeba_grouping}, for the Taskonomy-5 experiments of Table \ref{tab:taskonomy2} and CelebA-9 experiments of Table \ref{tab:celeba}, respectively. 

We note that in the prior state-of-the-art HOA \cite{hoa2020}, TAG \cite{tag2021}, and MTG-Net \cite{mtg2022}, a given task could be assigned to multiple groups. In that case, only the best performance of the specific group categorization, instead of the averaged performance across all categorizations, is reported as the final result. This imposes an additional requirement of a carefully collected validation set to determine the best categorization for tasks categorized into multiple groups. In contrast, our method formulates group categorization as the learning of a \texttt{Categorical} distribution, which ensures that each task is categorized \emph{exclusively and uniquely} into a single group. Consequently, our group categorization can be directly applied for testing, eliminating the need for a validation set to select the best categorization. In addition, the number of discovered groups may be less than the candidate input group number $K$ in our method, as our flexible \texttt{Categorical} distribution formulation allows each group to contain 0 to $N$ tasks, \ie, Eq. \eqref{eq:grouping}.

We also note that the group categorizations are not consistent \emph{w.r.t.} different targeting groups $K$, and this phenomenon appears for all the methods. For example, tasks A and B may be categorized into the same group given a certain $K$ but that categorization no longer holds when $K$ changes. One intuition behind this lies in the fact that the best task affinities, no matter the pairwise or high-order scenarios, are affected by different targeting groups $K$, especially when $K$ is significantly smaller than the number of input tasks $N$. 

\textbf{Intuitions About the Grouped Tasks.} As shown in Table \ref{tab:taskonomy_grouping}, our method has very similar grouping results as MTG-Net \cite{mtg2022} when $K=5$ for the Taskonomy-5 experiment, where the surface normal and the semantic segmentation are categorized into the same groups. Intuitively, this might imply that as a large-scale indoor scene dataset, \emph{most objects have planar surfaces (e.g., table desktop, wall, etc) in Taskonomy, therefore the surface normal (i.e., different planar surfaces) can be a good cue to identify different semantics (i.e., different objects)}. We further note that \emph{the categorization of different tasks can be dataset-dependent} (e.g., the surface normal may no longer hold a good cue for the semantic segmentation when most objects do have planar surfaces in another dataset), and \emph{our method is expected to well capture that dataset-dependent information through a fully data-driven manner}.



\begin{table*}[t]
\centering
\resizebox{0.99\textwidth}{!}{
    \begin{tabular}{c|l|c|ccccc}
    \hline
Groups&\makecell[c]{Method}                   & Categorizations  & Depth estimation  & Surface normal    & Semantic segmentation & Keypoint detection & Canny edge detection \\ \hline
\multirow{12}{*}{$K=3$}     &\multirow{3}{*}{HOA}     & group 1 & $\blacksquare$         & $\blacksquare$ & $\blacksquare$             &                   &                     \\
                         && group 2 &                  & $\square$         &                      & $\square$          &                     \\
                         && group 3 &                 &                  &                      & $\blacksquare$  & $\blacksquare$    \\ \cline{2-8}
&\multirow{3}{*}{TAG}     & group 1 & $\blacksquare$ &                   & $\blacksquare$     &                   &                     \\
                         && group 2 & $\square$         & $\blacksquare$ &                      &                   &                     \\
                         && group 3 &                  &                  &                      & $\blacksquare$  & $\blacksquare$    \\ \cline{2-8}
&\multirow{3}{*}{MTG-Net} & group 1 & $\square$         & $\blacksquare$ & $\blacksquare$     & $\square$          & $\square$            \\
                         && group 2 & $\square$         & $\square$         &                      & $\blacksquare$  &                     \\
                         && group 3 & $\blacksquare$ & $\square$         &                      &                   & $\blacksquare$    \\ \cline{2-8}
&\multirow{3}{*}{Ours}    & group 1 &                  &                  &                      &                   &                     \\
                         && group 2 & $\blacksquare$ &                  &                      &                   & $\blacksquare$    \\ 
                         && group 3 &                  & $\blacksquare$ & $\blacksquare$     & $\blacksquare$  &     \\ \hline \hline
\multirow{16}{*}{$K=4$}     &\multirow{4}{*}{HOA}     & group 1 &                  & $\square$         &                      & $\square$          &                     \\
                         && group 2 &                  &                  &                      & $\blacksquare$  & $\blacksquare$    \\
                         && group 3 &                  & $\blacksquare$ & $\blacksquare$     &          &            \\
                         && group 4 & $\blacksquare$ & $\square$         &                      &          &            \\ \cline{2-8}
&\multirow{4}{*}{TAG}     & group 1 & $\square$         &                  & $\square$             &                   &                     \\
                         && group 2 & $\blacksquare$ & $\square$         &                      &                   &                     \\
                         && group 3 &                  &                  &                      & $\blacksquare$  & $\blacksquare$    \\
                         && group 4 & $\square$         & $\blacksquare$ & $\blacksquare$     &          &            \\ \cline{2-8}
&\multirow{4}{*}{MTG-Net} & group 1 & $\square$         & $\blacksquare$ & $\blacksquare$     & $\square$          & $\square$            \\
                         && group 2 & $\square$         & $\square$         &                      & $\blacksquare$  &                     \\
                         && group 3 & $\blacksquare$ & $\square$         &                      &                   & $\blacksquare$    \\
                         && group 4 & $\square$         & $\square$         &                      &                   &                     \\ \cline{2-8}
&\multirow{4}{*}{Ours}    & group 1 & $\blacksquare$ &                  &                      &                   &     \\
                         && group 2 &         &                  &                      &                   &            \\
                         && group 3 &                  & $\blacksquare$ & $\blacksquare$     & $\blacksquare$  & $\blacksquare$                    \\
                         && group 4 &         &                  &                      &                   &  \\ \hline \hline
\multirow{20}{*}{$K=5$}     &\multirow{5}{*}{HOA}     & group 1 &                  & $\square$         &                      & $\square$          &                     \\
                         && group 2 &                  &                  &                      & $\blacksquare$  & $\blacksquare$    \\
                         && group 3 &                  & $\blacksquare$ & $\blacksquare$     &          &            \\
                         && group 4 & $\blacksquare$ & $\square$         &                      &          &            \\
                         && group 5 &                  & $\square$         &                      &          &            \\ \cline{2-8}
&\multirow{5}{*}{TAG}     & group 1 & $\square$         &                  & $\square$             &                   &                     \\
                         && group 2 & $\blacksquare$ & $\square$         &                      &                   &                     \\
                         && group 3 &                  &                  &                      & $\blacksquare$  & $\blacksquare$    \\
                         && group 4 & $\square$         & $\blacksquare$ & $\blacksquare$     &          &            \\
                         && group 5 & $\square$         & $\square$         & $\square$             & $\square$          &                      \\ \cline{2-8}
&\multirow{5}{*}{MTG-Net} & group 1 & $\square$         & $\blacksquare$ & $\blacksquare$     & $\square$          & $\square$            \\
                         && group 2 & $\square$         & $\square$         &                      & $\blacksquare$  &                     \\
                         && group 3 & $\blacksquare$ & $\square$         &                      &                   & $\blacksquare$    \\
                         && group 4 & $\square$         & $\square$         &                      &                   &                     \\
                         && group 5 &                  &                  &                      &                   &                     \\ \cline{2-8}
&\multirow{5}{*}{Ours}    & group 1 &          &                   &                       & $\blacksquare$  & \textcolor{red}{}            \\
                         && group 2 &         &                  &                      &                   & $\blacksquare$    \\
                         && group 3 & $\blacksquare$ &         &            &           &                     \\
                         && group 4 &         &                  &                      &                   &            \\
                         && group 5 &                  & $\blacksquare$ & $\blacksquare$     &                   &                     \\ \hline
\end{tabular}}
\caption{Categorization results on Taskonomy-5 experiments with $K = 3, 4, 5$. The parameters are identical to those in Table \ref{tab:taskonomy2}. Note that the prior state-of-the-art methods, \ie, HOA, TAG, and MTG-Net, may categorize a certain task into multiple groups. In that case, we count on the group with the best performance and report that in Table \ref{tab:taskonomy2}, which is denoted by the solid square $\blacksquare$. Otherwise, We denote by the hollow square $\square$. While each task is categorized \emph{exclusively and uniquely} into a single group in our method. We also note that by formulating MTG as the learning of a \texttt{Categorical} distribution, the flexibility of our method enables to finalize equal or less than $K$ groups, as we allow each group to contain 0 to $N$ tasks in Eq. \eqref{eq:grouping}.}
\label{tab:taskonomy_grouping}
\end{table*}

\begin{table*}[t]
\centering
\resizebox{0.99\textwidth}{!}{
    \begin{tabular}{c|l|c|ccccccccc}
    \hline
Groups&\makecell[c]{Methods}               & Categorizations  & 5\_o\_Clock\_Shadow & Black\_Hair & Blond\_Hair & Brown\_Hair & Goatee & Mustache & No\_Beard & Rosy\_Cheeks & Wearing\_Hat \\ \hline
\multirow{6}{*}{$K=2$}  &\multirow{2}{*}{HOA}  & group 1 & $\blacksquare$          & $\blacksquare$  & $\blacksquare$  & $\blacksquare$  &           & $\blacksquare$ &           & $\blacksquare$   & $\blacksquare$   \\
                      && group 2 &                    &            &            &            & $\blacksquare$ &           & $\blacksquare$ &             &             \\ \cline{2-12}
&\multirow{2}{*}{TAG}  & group 1 & $\blacksquare$          &            &            &            &           &           & $\blacksquare$ &             &             \\
                      && group 2 &                    & $\blacksquare$  & $\blacksquare$  & $\blacksquare$  & $\blacksquare$ & $\blacksquare$ &           & $\blacksquare$   & $\blacksquare$   \\ \cline{2-12}
&\multirow{2}{*}{Ours} & group 1 & $\blacksquare$                   & $\blacksquare$           &            & $\blacksquare$           &       &         &          &             & $\blacksquare$            \\
                      && group 2 &                    &            & $\blacksquare$           &            & $\blacksquare$      & $\blacksquare$        & $\blacksquare$         & $\blacksquare$            &             \\     \hline \hline
\multirow{9}{*}{$K=3$}  &\multirow{3}{*}{HOA}  & group 1 & $\blacksquare$          &            & $\blacksquare$  &            & $\blacksquare$ & $\blacksquare$ & $\square$         & $\blacksquare$   & $\blacksquare$   \\
                      && group 2 &                    & $\blacksquare$  &            & $\blacksquare$  &           &           &           &             &             \\
                      && group 3 &                    &            &            &            & $\square$         &           & $\blacksquare$ &             &             \\ \cline{2-12}
&\multirow{3}{*}{TAG}  & group 1 & $\blacksquare$          &            &            &            &           &           & $\square$         &             &             \\
                      && group 2 &                    & $\square$          & $\blacksquare$  &            & $\blacksquare$ & $\blacksquare$ & $\blacksquare$ & $\blacksquare$   & $\blacksquare$   \\
                      && group 3 &                    & $\blacksquare$  &            & $\blacksquare$  &           &           &           &             &             \\ \cline{2-12}
&\multirow{3}{*}{Ours} & group 1 & $\blacksquare$          &            &            & $\blacksquare$  & $\blacksquare$ &           & $\blacksquare$ &             & $\blacksquare$   \\
                      && group 2 &                    & $\blacksquare$  &            &            &           & $\blacksquare$ &           &             &             \\
                      && group 3 &                    &            & $\blacksquare$  &            &           &           &           & $\blacksquare$   &             \\     \hline \hline
\multirow{12}{*}{$K=4$}  &\multirow{4}{*}{HOA}  & group 1 & $\blacksquare$          &            &            &            & $\square$         &           &           &             &             \\
                      && group 2 &                    & $\square$          & $\blacksquare$  &            & $\blacksquare$ & $\blacksquare$ & $\square$         & $\blacksquare$   & $\blacksquare$   \\
                      && group 3 &                    & $\blacksquare$  &            & $\blacksquare$  &           &           &           &             &             \\
                      && group 4 &                    &            &            &            & $\square$         &           & $\blacksquare$ &             &             \\ \cline{2-12}
&\multirow{4}{*}{TAG}  & group 1 & $\blacksquare$          &            &            &            &           &           & $\blacksquare$ &             &             \\
                      && group 2 &                    & $\square$          & $\blacksquare$  & $\square$          &           &           & $\square$         & $\blacksquare$   & $\blacksquare$   \\
                      && group 3 &                    & $\blacksquare$  &            & $\blacksquare$  &           &           &           &             &             \\
                      && group 4 &                    &            &            &            & $\blacksquare$ & $\blacksquare$ &           &             &             \\ \cline{2-12}
&\multirow{4}{*}{Ours} & group 1 &                    &            &            &            &           &           &           &             & $\blacksquare$   \\
                      && group 2 & $\blacksquare$          &            & $\blacksquare$  &            &           & $\blacksquare$ &           &             &             \\
                      && group 3 &                    &            &            & $\blacksquare$  &           &           &           &             &             \\
                      && group 4 &                    & $\blacksquare$  &            &            & $\blacksquare$ &           & $\blacksquare$ & $\blacksquare$   &  \\     \hline
\end{tabular}}
    \caption{Categorization results on CelebA-9 experiments with $K = 2, 3, 4$. The parameters are identical to those in Table \ref{tab:celeba}. The solid square $\blacksquare$ represents the group with the best performance, and the hollow square $\square$ denotes the cases otherwise, when a certain task is categorized into multiple groups in HOA and TAG. Other conventions and explanations are the same as Table \ref{tab:taskonomy_grouping}.}
    \label{tab:celeba_grouping}
\end{table*}

\end{document}